\documentclass[lettersize,journal]{IEEEtran}
\usepackage{amsmath,amsfonts}
\usepackage{algorithmic}
\usepackage{array}
\usepackage[caption=false,font=normalsize,labelfont=sf,textfont=sf]{subfig}
\usepackage{textcomp}
\usepackage{stfloats}
\usepackage{url}
\usepackage{verbatim}
\usepackage{graphicx}
\def\BibTeX{{\rm B\kern-.05em{\sc i\kern-.025em b}\kern-.08em
    T\kern-.1667em\lower.7ex\hbox{E}\kern-.125emX}}
\usepackage{balance}

\usepackage{subfiles}
\usepackage{booktabs}
\usepackage{makecell}
\usepackage{bbding}
\usepackage{xcolor}
\usepackage[ruled]{algorithm2e}
\usepackage{floatrow}
\newfloatcommand{capbtabbox}{table}[][\FBwidth]

\usepackage{multirow}
\usepackage{subfig} 
\usepackage{subfloat}
\usepackage{hyperref}
\hypersetup{hidelinks,
	colorlinks=true,
	allcolors=black,
	pdfstartview=Fit,
	breaklinks=true}

\begin{document}
\title{Toward 3D Spatial Reasoning for Human-like Text-based Visual Question Answering}
\author{IEEE Publication Technology Department
\thanks{Manuscript created October, 2020; This work was developed by the IEEE Publication Technology Department. This work is distributed under the \LaTeX \ Project Public License (LPPL) ( http://www.latex-project.org/ ) version 1.3. A copy of the LPPL, version 1.3, is included in the base \LaTeX \ documentation of all distributions of \LaTeX \ released 2003/12/01 or later. The opinions expressed here are entirely that of the author. No warranty is expressed or implied. User assumes all risk.}}

\author{
Hao Li, Jinfa Huang, Peng Jin, Guoli Song,\\ Qi Wu, \IEEEmembership{Member, IEEE}, Jie Chen, \IEEEmembership{Member, IEEE}
\thanks{Hao Li, Jinfa Huang, Peng Jin are with the School of Electronic and Computer Engineering, Peking University, China (emails: lihao1984@pku.edu.cn, jinfahuang@stu.pku.edu.cn, jp21mails@gmail.com)} 
\thanks{Guoli Song and Jie Chen are with the Peng Cheng Laboratory, Shenzhen, China (emails: songgl@pcl.ac.cn, chenj@pcl.ac.cn)}
\thanks{Qi Wu is with the University of Adelaide, North Terrace, Adelaide SA 5005, Australia (email: qi.wu01@adelaide.edu.au)}
}

\markboth{IEEE TRANSACTIONS ON IMAGE PROCESSING}
{Author1, Author2, 
\MakeLowercase{\textit{(et al.)}: 
Paper Title}}

\maketitle

\begin{abstract}
Text-based Visual Question Answering~(TextVQA) aims to produce correct answers for given questions about the images with multiple scene texts. In most cases, the texts naturally attach to the surface of the objects. Therefore, spatial reasoning between texts and objects is crucial in TextVQA. However, existing approaches are constrained within 2D spatial information learned from the input images and rely on transformer-based architectures to reason implicitly during the fusion process. 
Under this setting, these 2D spatial reasoning approaches cannot distinguish the fine-grain spatial relations between visual objects and scene texts on the same image plane, thereby impairing the interpretability and performance of TextVQA models. 
In this paper, we introduce 3D geometric information into a human-like spatial reasoning process to capture the contextual knowledge of key objects step-by-step.
Specifically, (i)~we propose a relation prediction module for accurately locating the region of interest of critical objects; (ii)~we design a depth-aware attention calibration module for calibrating the OCR tokens' attention according to critical objects. Extensive experiments show that our method achieves state-of-the-art performance on TextVQA and ST-VQA datasets. More encouragingly, our model surpasses others by clear margins of 5.7\% and 12.1\% on questions that involve spatial reasoning in TextVQA and ST-VQA valid split. Besides, we also verify the generalizability of our model on the text-based image captioning task.
\end{abstract}

\begin{IEEEkeywords}
Text-based Visual Question Answering, Spatial Reasoning, 3D Geometric Information, Transformer
\end{IEEEkeywords}

\section{Introduction}

\IEEEPARstart{T}{asks} around scene-text have broad prospects in various areas including automatic driving and online shopping. As a representative task focusing on scene-text in images, text-based visual question answering (TextVQA)~\cite{mishra2019ocr,BitenTMBRJVK19,singh2019towards} needs to answer the given question about the image with multiple objects and scene texts. Thus, in addition to the simple object-object spatial reasoning in the VQA task~\cite{qiu2019incorporating},
TextVQA models also require the ability to read the textual information and reason the spatial relationship between different visual entities and texts in the image.

\begin{figure}[tbp]  
    \centering 
    \includegraphics[width=1.0\textwidth]{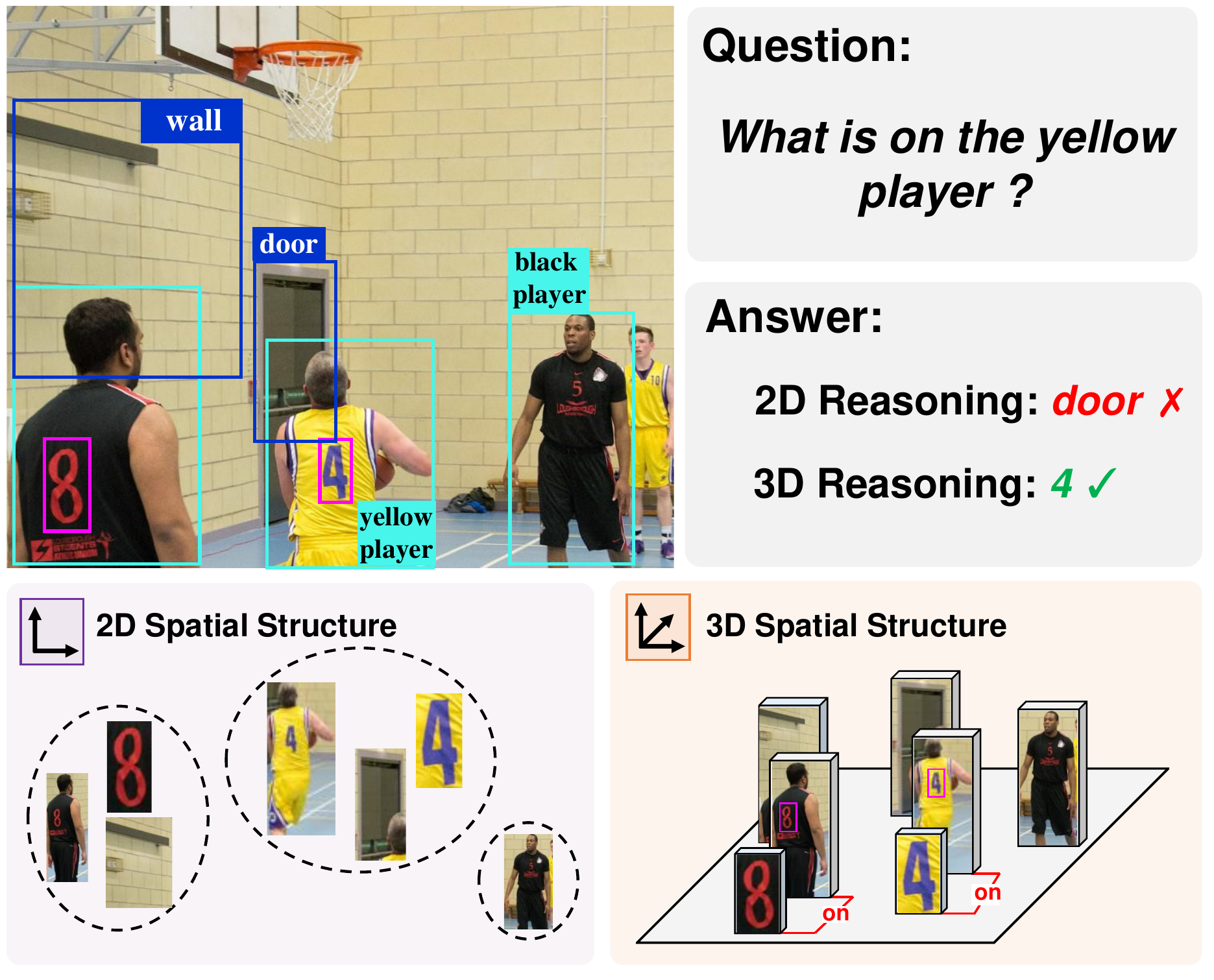} 
    \caption{
Illustration of the difference between spatial relation prediction of the 2D spatial reasoning method~(Right top) and our 3D spatial reasoning method~(Right bottom). The 2D reasoning methods can cluster objects and OCRs nearby, but cannot distinguish their fine-grained spatial relations. Our 3D reasoning method with depth information can correctly predict the fine-grained 3D spatial relations and is beneficial for spatial-related questions.
    }
    \label{motivation} 
    \vspace{-0.1in}
\end{figure}

\begin{figure}[tbp]  
    \centering 
    \includegraphics[width=1.0\textwidth]{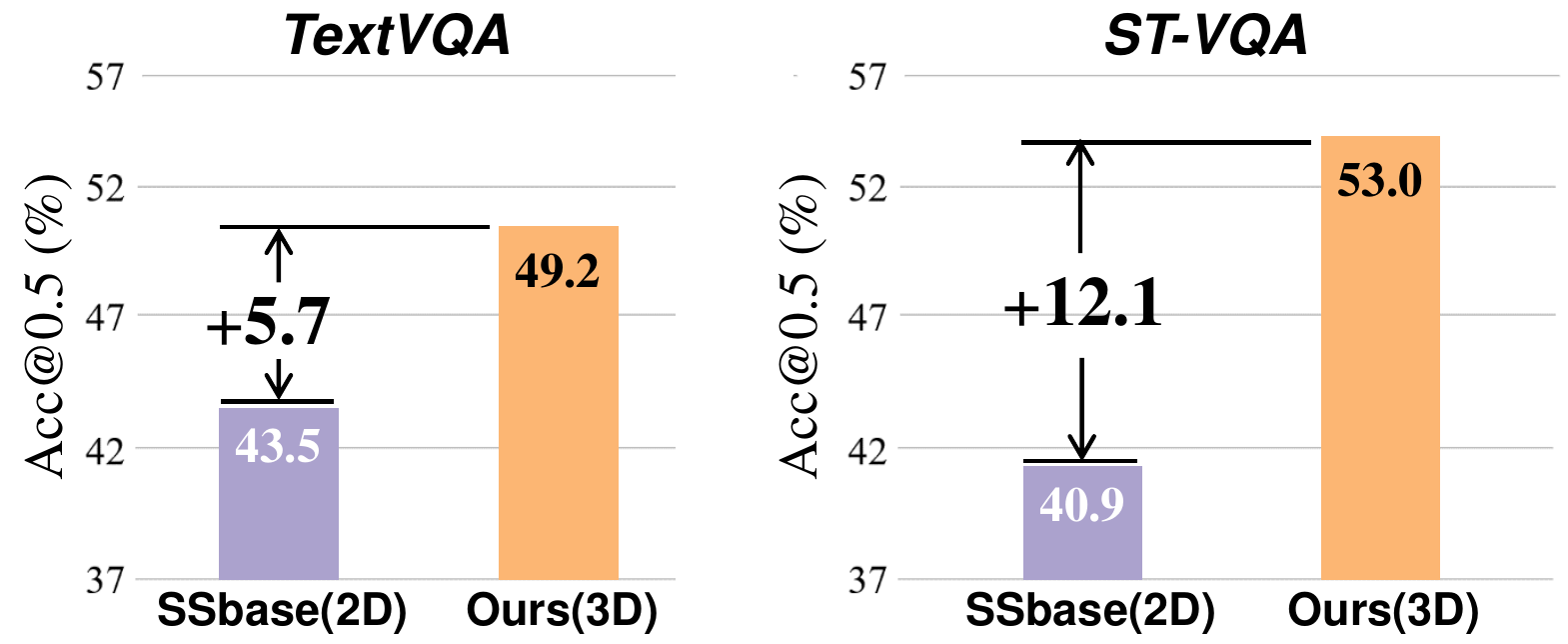} 
    \caption{
    Visualization of the performances of our 3D model and the 2D baseline model (SSbase). We split out all the spatial-related questions in TextVQA and STVQA datasets and create their 3D subsets. Compared with SSbase which applies 2D spatial reasoning, our 3D spatial reasoning method has a remarkable improvement on the 3D subsets from both datasets.
    }
    \label{performance} 
\end{figure}

Existing Transformer-based TextVQA models generally adopt the 2D spatial information learned from images for spatial reasoning. Most models~\cite{zeng2021beyond,zhu2020simple} use the absolute spatial information of objects and optical character recognition (OCR) tokens to supplement visual features. They straightly add the spatial features into the visual feature as the fusion of spatial and contextual information. Some inspiring works~\cite{gao2021structured,kant2020spatially} also explore rule-based spatial correlations from objects' and OCRs' absolute coordinates. Generally, they use the vanilla transformer structure to implicitly reason for the answer during the multimodal fusion process. However, without structured model input, the transformers only show shallow reasoning ability on the reasoning tasks~\cite{helwe2021reasoning}.

Therefore, existing TextVQA models cannot distinguish fine-grained spatial relations between multiple objects and scene-texts in the same image plane.
As shown in Fig.~\ref{motivation}, the 2D spatial reasoning model receives the bounding box information, which only contains 2D spatial information.  It leads to spuriousness when multiple object bounding boxes and OCR bounding boxes are close to each other.  
To this end, the model cannot distinguish the spatial relations between ``4'', ``door'', and ``yellow player''.  Due to the wrong prediction that ``door'' and ``4'' are on the same object ``yellow player'', ``door'' becomes the distraction leading the model to predict wrong. 

\begin{figure}[tbp]  
\vspace{-0.1in}
    \centering 
    \includegraphics[width=1.0\textwidth]{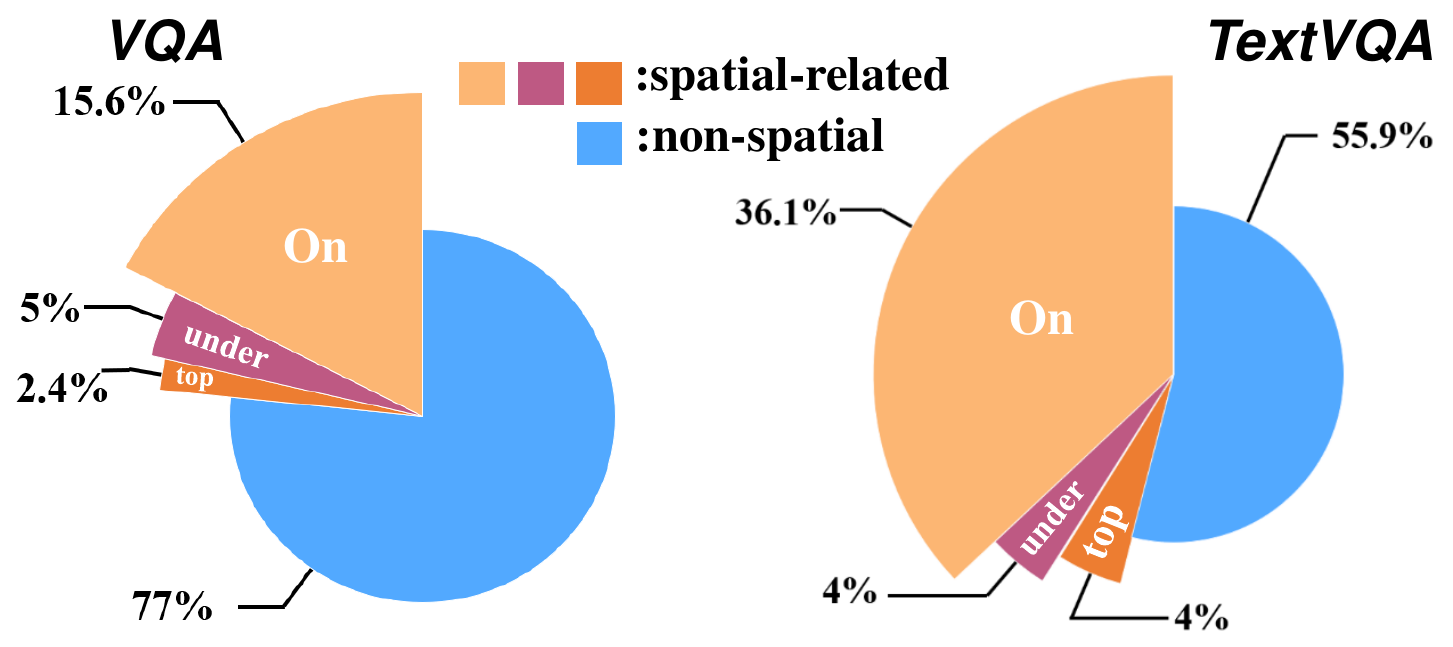}
    \caption{
    The analysis of the proportion of the spatial-related question in the most popular datasets for VQA and TextVQA tasks (Visual Genome~\cite{krishna2017visual} and TextVQA~\cite{DBLP:conf/cvpr/SinghNSJCBPR19}). The three colors in the spatial-related part represent questions containing ``on, under, top''. The pie chart shows that spatial-related questions in TextVQA are far more than those in VQA (44.1\% vs 23.0\%), indicating the importance of spatial reasoning ability in the TextVQA task.
    }
    \label{figure3} 
    \vspace{-0.1in}
\end{figure}


Fig.~\ref{figure3} further shows the importance of improving the spatial reasoning ability of the  model for the TextVQA task. 
Surprisingly, spatial-related questions account for 44.1\% of the total TextVQA dataset. In contrast, only 23.0\% of the questions in the standard VQA dataset use spatial prepositions. 
Hence, to achieve the complex spatial reasoning on multiple objects and OCR tokens, 
we introduce 3D geometric information into TextVQA reasoning model.
The adoption of 3D geometric information is motivated by an intuitive observation: in most cases, the texts attach to the surface of objects. Moreover, with the depth information, the model can better understand the real-world 3D scene inside the images~\cite{banerjee2021weakly}. Therefore, the 3D geometric information might assist the TextVQA model to understand the spatial structure of visual concepts in an image,
which is extremely important for the reasoning process of transformers~\cite{vieu1997spatial,chang2014learning}.
As shown in Fig.~\ref{motivation}, with the guidance of 3D  information, our model alleviates the spuriousness problem on the same image plane and accurately predicts the fine-grained spatial relations such as ``4-on-yellow player'' and ``door-behind-yellow player''.  Benefiting from correct spatial relation prediction, our model answers the question correctly.


However, it is challenging for TextVQA models to take advantage of 3D geometric information. Previous models apply the vanilla transformer structure as the fusion module, but this structure lacks spatial inductive biases and hardly preserves 3D spatial structure of the input image~\cite{zhou2022sp,wu2021cvt,yuan2021incorporating,yuan2021tokens}.
Since existing models are delicately designed for 2D spatial reasoning, they cannot comprehend 3D geometric information by directly replacing 2D spatial information.

In this paper, to facilitate 3D understanding ability of the model, we propose the Depth-Aware TextVQA Network (DA-Net) with two novel modules: the depth-aware attention calibration module and the relation prediction module.
Our depth-aware attention calibration module follows a step-by-step knowledge transferring procedure: it first generates attentions for object tokens, and then transfers the object attention score for calibrating the OCR tokens' attention under the guidance of 3D geometric information. Such a step-by-step process is similar to human reasoning~\cite{vieu1997spatial}: humans first locate the critical objects in the image based on the given question; then they identify the OCR tokens on the detected objects and get the correct answer. 
To model the above gradual working patterns with 3D geometric information, our DA-Net introduces a relation prediction module to strengthen the model's understanding of 3D spatial relationships, which is beneficial for locating the critical objects.
 
To demonstrate the effectiveness of our DA-Net, we conduct the experiments on two widely-used datasets, the TextVQA dataset~\cite{DBLP:conf/cvpr/SinghNSJCBPR19} and the ST-VQA dataset~\cite{BitenTMBRJVK19}. Extensive experiments demonstrate that our method outperforms the state-of-the-art TextVQA methods. 
We also split the spatial-related questions in both datasets and generate their 3D subsets. Fig.~\ref{performance} shows that our model achieves the state-of-the-art performance on the TextVQA task and has a remarkable improvement compared with the baseline model: SSbase (e.g., 5.7\% on the TextVQA 3D subset and 12.1\% on the ST-VQA 3D subset). 
The visualization for spatial maps and attention scores further prove our DA-Net can understand and leverage 3D spatial information. What's more, we achieve promising performance on the text-based image captioning task, which verifies the generalization of our model.

Overall, the main contributions of our work are as follows:
\begin{itemize} 
\item As far as we know, we are the first to introduce 3D geometric information into the TextVQA task, in order to effectively handle the spuriousness of existing 2D models' spatial reasoning process.

\item 
With 3D spatial information, we propose a relation prediction task to strengthen the model's understanding of 3D spatial relationship for better locating the region of interest of key objects. 

\item To identify the OCR tokens on the corresponding object, we design a depth-aware attention calibration module to calibrate the OCR token attention based on the key object attention score.

\item 
Extensive experiments show that our model achieves the state-of-the-art performance on two TextVQA datasets. Results on
their 3D subsets further demonstrate the spatial understanding and reasoning ability of our model.

\end{itemize}

\section{Related work}
\subsection{Text-based Visual Question Answering} 
As a further task of VQA~\cite{li2022joint,johnson2017clevr,cao2019interpretable}, TextVQA has seen rapid development in recent years. The TextVQA task aims to understand and reason scene texts in images. A model first detects OCR text and visual objects in the images and then answers related questions. LoRRA~\cite{singh2019towards} extends Pythia~\cite{jiang2018pythia} with an OCR detection module and a cross-modality attention module to reason over a combined list of answers from a static vocabulary and detected OCR tokens. M4C~\cite{hu2020iterative} replaces the copy mechanism in LoRRA with a dynamic pointer network and utilizes a transformer to combine multimodal information into a joint embedding space. SSbaseline~\cite{zhu2020simple} claims a simple attention mechanism can obtain comparable performance with previous sophisticated multi-modality frameworks.
Recent works~\cite{chen2020uniter,li2020oscar,lu2019vilbert,yang2021tap} apply large-scale pretraining datasets on TextVQA.
However, these existing models straightly add a location embedding of the absolute location to the object and OCR features, lacking explicit spatial reasoning between different visual contents.

\subsection{Leveraging Spatial Reasoning for TextVQA} 
Spatial reasoning ability is significant in behavioristics and artificial intelligence~\cite{wang2021improving,liu2021learning}.
In the visual reasoning domain, including VQA~\cite{wu2017visual} and VideoQA \cite{sun2021video} tasks, the region-level spatial relationship has been proved significantly helpful~\cite{li2019relation,yao2018exploring,narasimhan2018out,ryoo2021tokenlearner,yang2020trrnet}. For TextVQA, MM-GNN~\cite{gao2020multi} represents an image as a graph consisting of three sub-graphs depicting visual, semantic, and numeric modalities. Then MM-GNN uses three aggregators to guide the reasoning process. SMA~\cite{gao2021structured} uses a structural graph representation to encode the object-object, object-text, and text-text relationships appearing in the image and then designs a multimodal graph attention network to reason over it. SA-M4C~\cite{kant2020spatially} proposes a novel spatially aware self-attention layer such that each visual entity only looks at neighboring entities defined by a spatial graph. However, these models extract the spatial relationship between different visual content from 2D bounding boxes. When extracting the crucial spatial relations for the TextVQA, such as the object-text spatial relationship, lacking 3D geometric information reduces the relation accuracy.

\subsection{Leveraging depth estimation in multimedia} 
Monocular (single-image) depth estimation remains a challenging problem, with learning-based methods pushing the envelope~\cite{saxena2005learning,eigen2014depth,li2017two}. AdaBins~\cite{bhat2021adabins} uses a transformer-based architecture that adaptively divides depth ranges into variable-sized bins and estimates depth as a linear combination of these depth bins. It is the state-of-the-art monocular depth estimation model for outdoor and indoor scenes. We use it as the depth extractor to guide our models for better spatial reasoning. In multiple areas, only using 2D images cannot adequately guide the model with the 3D geometry~\cite{saxena2005learning}. Thus, the depth estimation method is leveraged across multimedia fields. For generation tasks, DaGAN~\cite{hong2022depth} uses the learned depth maps to estimate sparse facial key points and generate highly realistic faces. DepthGAN~\cite{shi20223d} uses depth maps as 3D prior to assist in synthesizing indoor scenes. For classification tasks such as object detection and visual question answering, pseudo-LiDAR-based methods~\cite{ma2019accurate,wang2019pseudo,weng2019monocular,li2022locality} lift images to 3D coordinate via monocular depth estimation. In \cite{banerjee2021weakly}, extra depth information is used as weak supervision to enhance the spatial estimation between objects. However, no existing models consider the depth correlation between objects and OCRs in TextVQA.

\subsection{2D Spatial reasoning VS 3D Spatial reasoning}
2D and 3D spatial reasoning occur in many computer vision and multimedia tasks including 
retrieval~\cite{jin2023diffusionret}, visual grounding~\cite{liu2021refer,cheng2023parallel}, VQA~\cite{ye20213d} and VideoQA \cite{sun2021video,li2023tg}.
In many scenarios, 2D visual information has intrinsic limitations such as illumination, pose, expression, and disguise~\cite{zhou20183d}. Containing richer geometric structures, 3D visual information can provide more discriminative spatial features for the spatial reasoning process and overcome the spuriousness of visual changes. Although 3D~(RGB-D) contains more abundant information than 2D~(RGB), there are still challenges to applying 3D information in the spatial reasoning process. 
Firstly, reasoning models need to design delicately to take advantage of 3D geometric information. Traditional 2D reasoning models hardly learn 3D geometric information by directly replacing the 2D spatial information.
Secondly, the acquisition of 3D information cannot be accomplished by crawling the Web like how 2D images are collected. 
For the TextVQA task, we propose to apply 3D spatial reasoning and also need to address the above two challenges. 
Compared with other tasks which only need to reason between different objects, TextVQA  requires spatial reasoning on multiple objects and OCR tokens.
To explore the geometric structure, we design a depth-aware attention calibration module and the relation prediction auxiliary task.
To obtain 3D annotations cheaply, we predict the 3D information using a well-pretrained depth estimation model instead of sampling by the depth camera.

\label{sec:related work}

\begin{figure*}[ht] 
    \vspace{-0.1in}
    \centering 
    \includegraphics[width=1\textwidth]{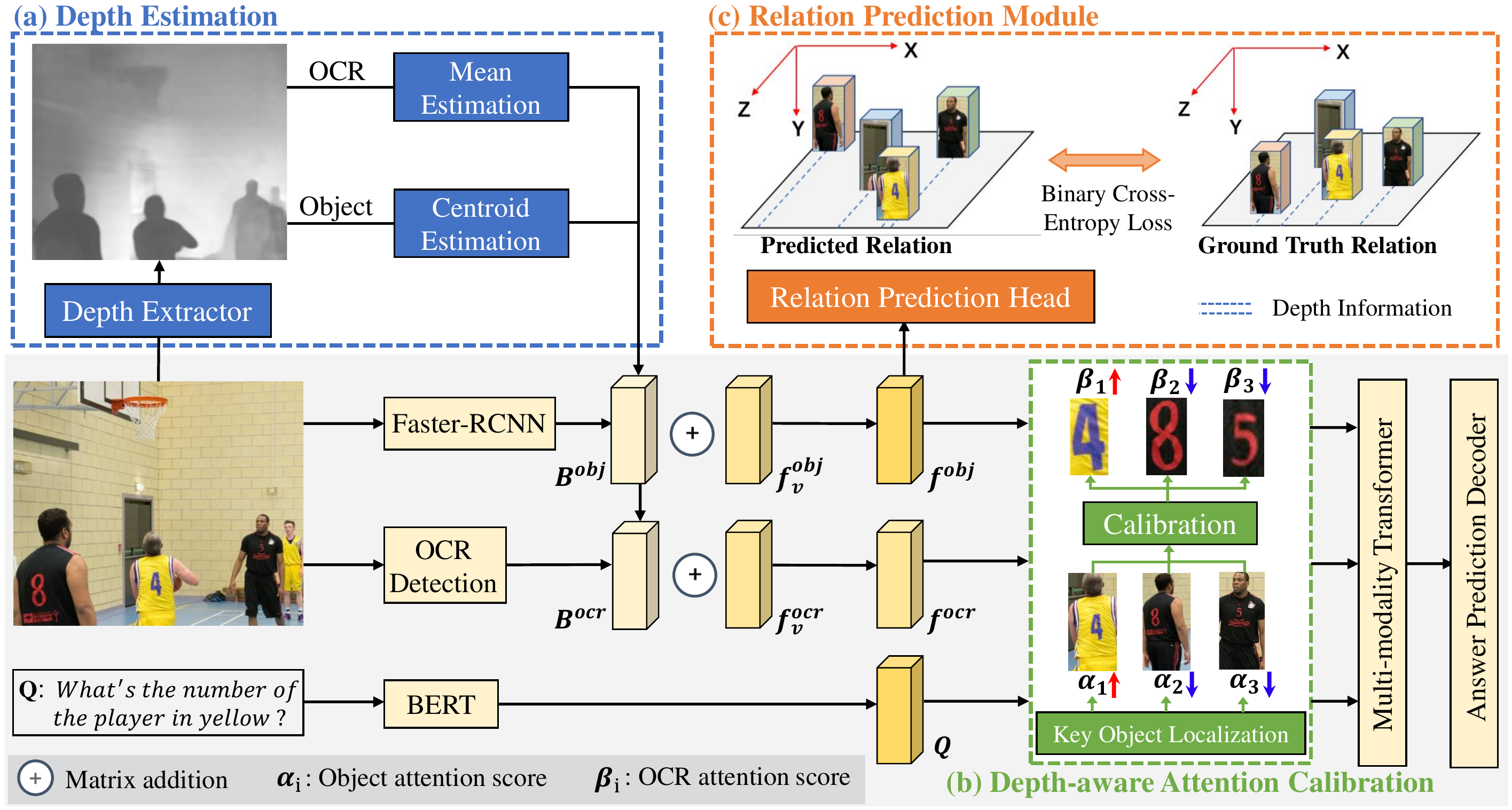} 
    \caption{
    An illustration of the proposed Depth-Aware TextVQA Network (DA-Net). It contains three modules to exploit 3D information. (a)~The depth estimation module extracts the pixel-level depth map and uses mean and centroid estimations to calculate the objects' and OCRs' depth. (b)~The depth-aware attention calibration module enhances the feature attention between depth-relevant OCR and key object tokens and calibrates the irrelevant OCR attention score. (c)~The relation prediction module predicts the image's 3D geometric structure using objects' features. In this part, the binary cross-entropy loss between predicted relation and GT relation helps our model understand 3D information.} 
    \label{overview} 
    \vspace{-0.1in}
\end{figure*}

\section{Depth-Aware TextVQA Network}

We propose the Depth-Aware TextVQA Network~(DA-Net) for the TextVQA task, applying the 3D geometric information in the spatial reasoning procedure. Fig.~\ref{overview} shows the pipeline of our DA-Net. Compared with the vanilla reasoning pipeline of TextVQA, we introduce three additional new modules.
(a)~Depth Estimation module extracts the relative depth map from the raw image and estimates the depth information for each OCR and object token. (b)~Depth-aware Attention Calibration module simulates the step-by-step human-like reasoning process. It also enhances the feature attention between depth-relevant OCR and object tokens. (c)~Relation Prediction Head further helps our model understand the 3D depth information and learn the implicit spatial structure from the image. We introduce the other components of our DA-Net in Section~\ref{Preliminaries}.

\vspace{-0.1in}
\subsection{Preliminaries}
\label{Preliminaries}
General TextVQA models such as \cite{hu2020iterative,han2020finding,wang2022tag} mainly consist of four components: multi-modality inputs,  OCR noise reduction, multi-modality transformer, and answer prediction decoder. Our DA-Net adopts the same structure as the general models for these components. 

\textbf{Multi-modality inputs.}\quad Given a text-related question and an image, models extract three modalities and prepare their corresponding features. Concretely, for question tokens, models use pre-trained BERT~\cite{devlin2018bert} to generate the question word embedding $Q = \{q_i\}_{i=1}^L$, which is $768\times L$ dimension and $L$ is the length of the question.
For OCR features, models use off-the-shelf OCR detection models~\cite{liu2019omnidirectional,wang2019simple} to locate and extract OCRs. For each image, we pad the OCR numbers into 50.
For object visual features, models use pre-trained object detectors~\cite{ren2015faster} to capture visual objects' positions and features. For each image, we pad the object numbers into 100.

\textbf{OCR Noise Reduction. }\quad The OCR tokens extracted by OCR detectors contain distractions including repeating OCR tokens and OCR subsequence tokens. Following SA-M4C~\cite{kant2020spatially} and LOGOs~\cite{lu2021localize}, we use denoising strategies to remove the distractions in OCR tokens. Specifically, we first calculate the IoU between each OCR token pair and remove the item with a high IoU score, which might be repeating OCR tokens. Then, we compare each OCR token pair and remove short subsequence tokens. 

\textbf{Multi-modality transformer.}\quad Most TextVQA models use the transformer as the fusion encoder of three modality inputs. For a fair comparison, we use the generative transformer~\cite{devlin2018bert} as our cross-modality fusion module to fuse $768$ dimension vectors from OCR, object, and question.

\textbf{Answer prediction decoder.}\quad M4C~\cite{hu2020iterative} proposes a powerful answer decoder module that iteratively decodes answers using a dynamic pointer network. Following other recent works~\cite{zhu2020simple,zeng2021beyond,gao2021structured}, our model applies the same answer prediction decoder as M4C for a fair comparison. 

\vspace{-0.1in}
\subsection{Depth Estimation}

\textbf{Pixel-level Depth Computation.}\quad To extract pixel-level depth information from images in TextVQA datasets~\cite{BitenTMBRJVK19,DBLP:conf/cvpr/SinghNSJCBPR19}, we utilize an open-source monocular depth computation method AdaBins~\cite{bhat2021adabins}, which is the state-of-the-art method. AdaBins divides the depth range into bins whose center value is estimated adaptively per image. The final depth values are estimated as linear combinations of the bin centers. We obtain depth-value $d(i, j)$ for each pixel $(i, j), i\in\{1,..., H\}, j\in\{1,..., W\}$ in the image, where $H$ is the height of the image  and $W$ is the width of the image.

\textbf{Extracting OCR depth using mean estimation. }
TextVQA models often use the bounding box for each OCR token in the image as its spatial information. Given an OCR token bounding box $[(x_1, y_1), (x_2, y_2)]$, $(x_1, y_1)$ and $(x_2, y_2)$ are the bounding box's top-left corner and bottom-right corner. The depth of the OCR token is calculated as the mean depth of all points in the bounding box:
\begin{gather}
    d^{ocr} = \frac{\sum_{i, j} d(i, j)}{|x_2-x_1|*|y_2-y_1|},\quad i\in [x_1, x_2],\quad j \in [y_1, y_2].
\end{gather}
Thus our model uses 3D bounding box $[B^{ocr}_i, d^{ocr}_{i}]$ as OCR token's spatial coordinates. $B^{ocr}_i$ is $OCR_i$'s bounding box.

\textbf{Extracting object depth using centroid estimation. }
Same as OCR tokens, general TextVQA models use the object bounding box $[(x_1, y_1), (x_2, y_2)]$ as the object's spatial information. However, most objects have irregular shapes, and their bounding boxes contain surrounding visual regions. We use centroid estimation to calculate objects' depths to avoid depth noises from surrounding regions. 
We define the object's centroid as $(x_c, y_c)$ and take 0.1 of the width and height of the bounding box to obtain the surrounding area $(\varepsilon_x, \varepsilon_y)$,
\begin{gather}
    x_c = \frac{x_1 + x_2}{2},\ y_c = \frac{y_1 + y_2}{2},\ \varepsilon_x = \frac{x_2-x_1}{20},\ \varepsilon_y=\frac{y_2-y_1}{20}.
\end{gather}
The depth of the object token is calculated as the mean depth of the area surrounding its centroid:
\vspace{-0.5em}
\begin{gather}
    d^{obj} = mean(d(i, j)),\\
    i\in [x_c - \varepsilon_x, x_c + \varepsilon_x],\quad j\in [y_c - \varepsilon_y, y_c + \varepsilon_y].  \nonumber
\end{gather}
Thus our model uses 3D bounding box $[B^{obj}_i, d^{obj}_{i}]$ as each object token's spatial coordinates. $B^{obj}_i$ is the bounding box of the $i^{th}$ object $obj_i$.

\vspace{-0.1in}
\subsection{Depth-aware Attention Calibration Module}

The multi-modality fusion part of TextVQA models needs to fuse homogeneous entities from OCRs, objects, and questions into a joint embedding space. However, it consumes much computation and cannot efficiently extract key token features from all entities.
To filter out irrelevant tokens and highlight key tokens, models~\cite{zeng2021beyond,zhu2020simple} use attention blocks to summarize token features from objects and OCRs. However, the previous summary module pays little attention to the 3D relation between OCRs and objects.

\begin{figure}[t] 
    \centering 
    \includegraphics[width=1.0\textwidth]{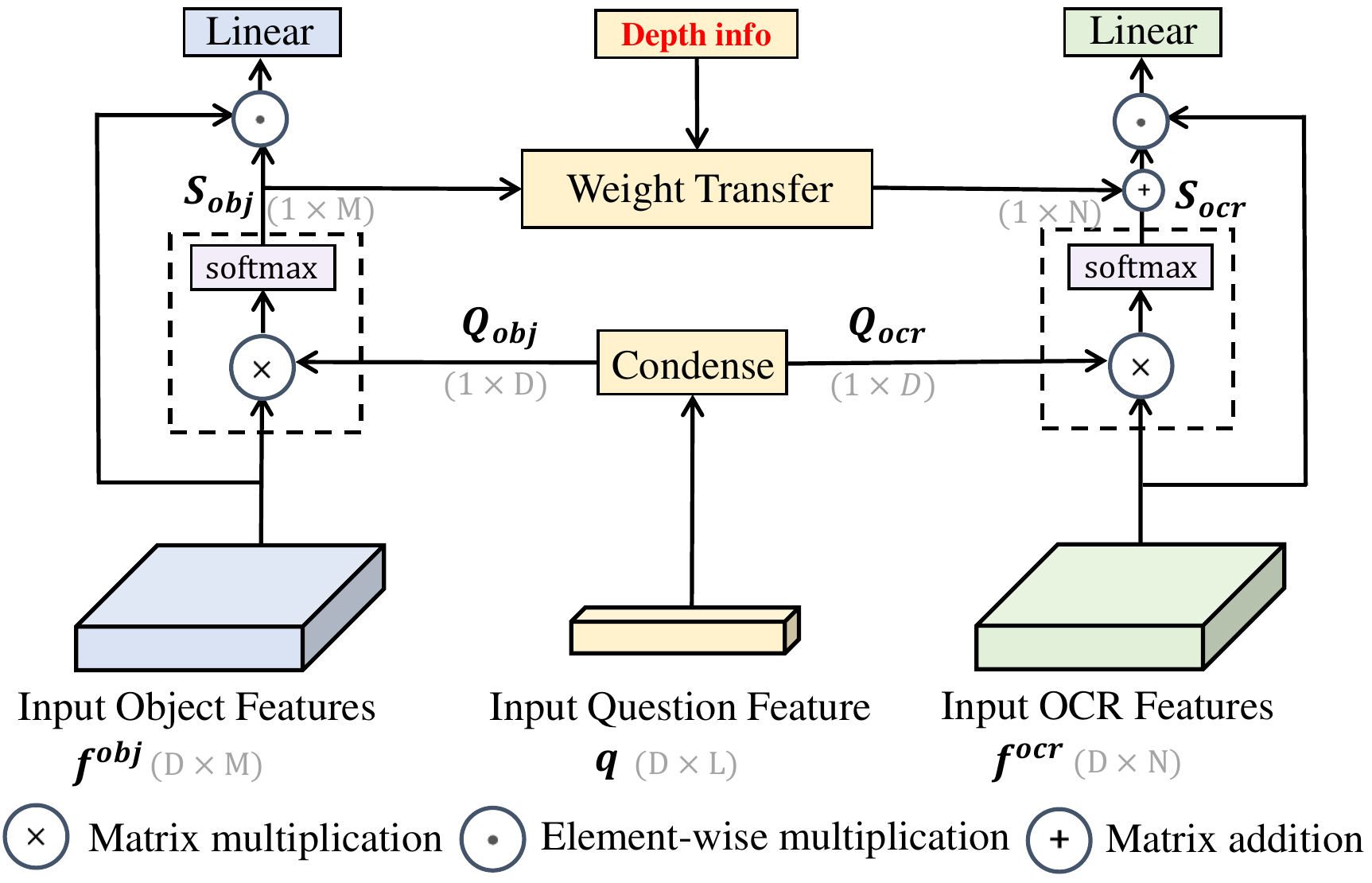} 
    \caption{Our depth-aware attention calibration module.} 
    \label{module_depth} 
\end{figure}

In this paper, we extend the feature summary module to Depth-aware Attention Calibration~(DAC) module. As shown  in Fig.~\ref{module_depth}, it uses depth-aware weight transfer over the input OCR and object tokens. The details are as follows.

\textbf{Condensing Question Feature. }\quad We start with the OCR tokens. To efficiently calculate the attention scores between OCR and the question sequence, we first condense the question sequence into an individual entity. Specifically, given a question word sequence $Q = \{q_i\}_{i=1}^L$, we use two convolution layers and one ReLU activation function between them to condense the question sequence into an individual entity $Q_{ocr}$.
\begin{gather}
    q_i^{ocr} = conv\{ReLU[conv(q_i)]\},\quad i\in \{1,...,L\};\\
    Q_{ocr} = \sum\nolimits_{i=1}^{L} q_i * Softmax(q_i^{ocr}).
\end{gather}
The $Q_{obj}$ for object features can be calculated in the same way.
\begin{gather}
    q_i^{obj} = conv\{ReLU[conv(q_i)]\},\quad i\in \{1,...,L\};\\
    Q_{obj} = \sum\nolimits_{i=1}^{L} q_i * Softmax(q_i^{obj}).
\end{gather}

\textbf{Attention Scores. }\quad We define object features $f^{obj} = (x_1,...,x_N) \in \mathbb{R}^{D\times N}$ and OCR features $f^{ocr} = (y_1, ... , y_M) \in \mathbb{R}^{D\times M}$. We use $S_{obj} = [\alpha_1, ... , \alpha_N]$ to represent the object attention score and $S_{ocr} = [\beta_1, ... , \beta_M]$ to represent the OCR attention score, where: 
\begin{gather}
    \alpha_i = Softmax(\frac{Q_{obj}(x_i)^T}{\sqrt{d}}),\quad \forall i \in \{1,...,N\},\\
    \beta_j = Softmax(\frac{Q_{ocr}(y_j)^T}{\sqrt{d}}),\quad \forall j \in \{1,...,M\}.
\end{gather}

\begin{figure}[t] 
    \vspace{-0.1in}
    \centering 
    \includegraphics[width=1.0\textwidth]{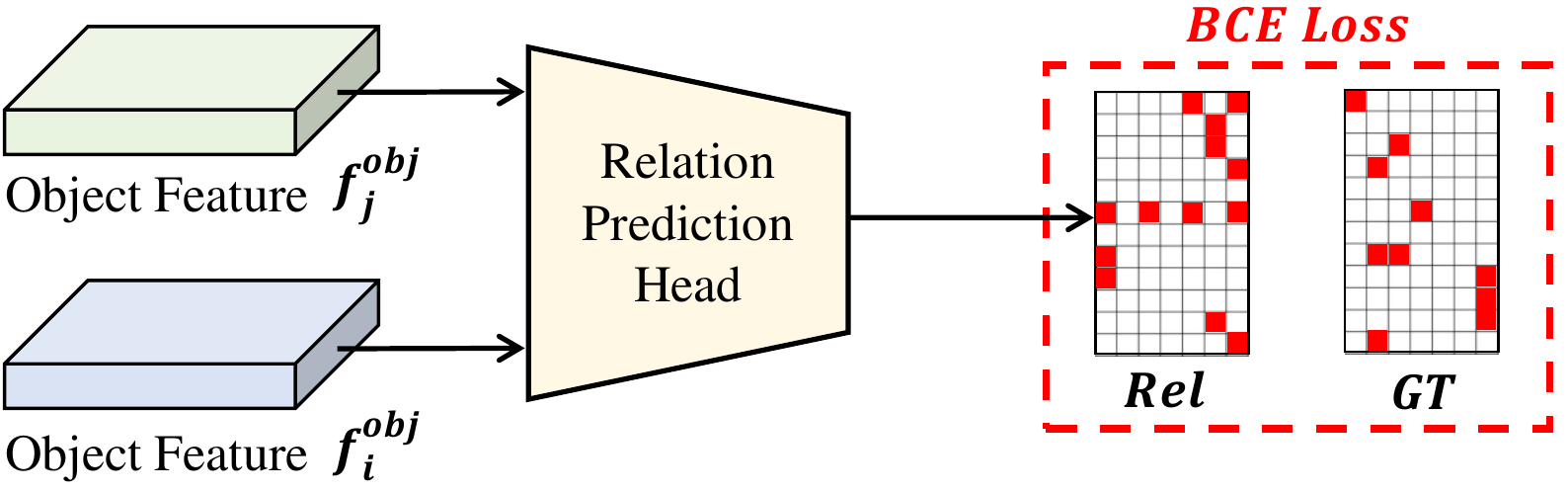} 
    \caption{The overview of our relation prediction module. We apply two linear layers as the relation prediction head. The predicted relation matrix $Rel$ calculates Binary Cross-Entropy (BCE) loss with the ground truth relation matrix $GT$.} 
    \label{module_loss} 
\end{figure}

\textbf{Depth-aware Weight Transfer. }\quad
To highlight the spatial relationship between OCRs and objects, we transfer object weights to OCR weights under the supervision of 3D spatial information. Firstly, we define $CR$ as the cover rate between the OCR bounding box $B^{ocr}$ and object bounding box $B^{obj}$. 
\begin{gather}
    CR(B^{obj}_{i}, B^{ocr}_{j}) = \frac{Area(B^{obj}_{i}\cap B^{ocr}_{j})}{Area(B^{ocr}_{j})}.
\end{gather}

Secondly, we generate the $\Delta_{ij}$ which represents the weight transfer rate from $obj_i$ to $ocr_j$. $\Delta_{ij}$ is positively correlated with the cover rate between the OCR bounding box and the object bounding box.
\begin{gather}
    \Delta_{ij} = Softmax(CR(B^{obj}_{i}, B^{ocr}_{j})*[1-(d^{obj}_{i}-d^{ocr}_{j})]).
\end{gather}

Thirdly, the attention score of $ocr_j$ in $S_{ocr}$ is added by the attention score of $obj_i$ in $S_{obj}$ with the weight transfer rate $\Delta_{ij}$. We compute the weighted sum of $f^{obj}$ and $f^{ocr}$ using the updated $S_{obj}$ and $S_{ocr}$. $F^{obj}$ and $F^{ocr}$ represent the summary feature for objects and OCRs. 
\begin{gather}
    F^{obj} = \sum S_{obj}\cdot f^{obj},\\
    F^{ocr} = \sum (S_{ocr}+\Delta\cdot S_{obj})\cdot f^{ocr}.
\end{gather}

\begin{table*}
    \begin{center}
    \caption{Acc(\%) on TextVQA dataset. In this paper, the \textcolor{red}{Red} colored numbers denote the best results across all non-pretrained approaches in Table. The \textcolor{blue}{Blue} colored numbers denote the second best results. }
    \label{table:TextVQA}
    \begin{tabular}{c|clcccc|ccc}
    
    \toprule
    Data \& Time & \# & Method & \makecell{OCR \\ system} & \makecell{Acc. \\ on val} & \makecell{Acc. \\ on test} & \makecell{Acc. on\\ 3D Subset} & \makecell{with\\STVQA}&\makecell{Acc. \\ on val} & \makecell{Acc. \\ on test}\\
    \midrule
    \multirow{3}*{\shortstack{1.4M Data\\ 520 GPU hours}} &\multicolumn{4}{l}{\emph{\textbf{Pretrained Models}}}&&&&&\\
    &\small\texttt{1} & LOGOs~\cite{lu2021localize} & Multiple-OCR & 51.5 & 51.1 &- &\Checkmark&51.53&51.08\\
    &\small\texttt{2} & TAP~\cite{yang2021tap} & MicroOCR & 54.7 & 53.9&- &\Checkmark&50.57&50.71\\
    \midrule
    \multirow{14}*{\shortstack{28K Data\\ 26 GPU hours}} &\multicolumn{5}{l}{\emph{\textbf{Non-Pretrained Models}}}&&&&\\
    &\small\texttt{3} & LoRRA~\cite{singh2019towards}& Rosetta-ml & 26.5 & 27.6 &- &-&-&-\\
    &\small\texttt{4} & {MM-GNN~\cite{gao2020multi}} & Rosetta-en & 31.4 & 31.1 &29.7&-&-&-\\
    &\small\texttt{5} & {M4C~\cite{hu2020iterative}} & Rosetta-en & 39.4 & 39.0 & 42.3 &\Checkmark&40.60&40.50\\
    &\small\texttt{6} & LaAP-Net~\cite{han2020finding} & Rosetta-en & 40.7 & 40.5 & - &\Checkmark&41.00&41.40\\
    &\small\texttt{7} & SMA~\cite{gao2021structured} & Rosetta-en & 40.0 & 40.6  & - &\Checkmark&44.60&45.50 \\
    &\small\texttt{8} & TAP~\cite{yang2021tap} & Rosetta-en & 44.1 & - & - & - & - & - \\
    &\small\texttt{9} & CRN~\cite{liu2020cascade} & Rosetta-en & 40.4 & 41.0 &- &-&-&-\\
    &\small\texttt{10} & PAT~\cite{zhang2021position} & Rosetta-en & 42.8 & 43.4 &- &-&-&-\\
    &\small\texttt{11} & {SA-M4C~\cite{kant2020spatially}}& Google-OCR & 45.4 & 44.6 &45.5 &\Checkmark&45.40&44.60\\
    &\small\texttt{12} & {SSbaseline~\cite{zhu2020simple}} & SBD-Trans & 43.9 & 44.7 & 43.5 &\Checkmark&45.53&45.66\\
    &\small\texttt{13} & BOV~\cite{zeng2021beyond} & SBD-Trans & 44.8 & \textcolor{blue}{\textbf{45.6}} &- &\Checkmark&\textcolor{blue}{\textbf{46.24}}&\textcolor{blue}{\textbf{46.96}}\\
    &\small\texttt{14} & \textbf{DA-Net (Ours)} & Rosetta-en & \textcolor{blue}{\textbf{46.8}} & 44.3 & \textcolor{blue}{\textbf{45.7}} & -& -& -\\
    &\small\texttt{15} & \textbf{DA-Net (Ours)} & SBD-Trans & \textcolor{red}{\textbf{47.2}} & \textcolor{red}{\textbf{46.6}} & \textcolor{red}{\textbf{49.2}} &\Checkmark& \textcolor{red}{\textbf{47.12}}&\textcolor{red}{\textbf{47.11}}\\
    \bottomrule
    
    \end{tabular}
    \end{center}
\end{table*}

\begin{table*}
    \begin{center}
    \vspace{-0.1in}
    \caption{Acc(\%) and ANLS(\%) on STVQA dataset. Higher is better in all columns. In this paper, the \textcolor{red}{Red} colored numbers denote the best results across all non-pretrained approaches in Table. The \textcolor{blue}{Blue} colored numbers denote the second best results. }
    \label{table:STVQA}
    \begin{tabular}{c|clccccc}
    
    \toprule
    Data \& Time & \# & Method & \makecell{OCR \\ system} & \makecell{Acc.(\%) \\ on val} & \makecell{ANLS (\%) \\ on val} & \makecell{ANLS (\%) \\ on test}& \makecell{Acc. \\ on 3D Subset}\\
    \toprule
    \multirow{3}*{\shortstack{1.4M Data\\ 520 GPU hours}} & \multicolumn{4}{l}{\emph{\textbf{Pretrained Models}}}&&&\\
    &\small\texttt{1} & LOGOs~\cite{lu2021localize} & Multiple-OCR & 48.6 & 58.1 & 57.9&-\\
    &\small\texttt{2} & TAP~\cite{yang2021tap} & MicroOCR & 50.8 & 59.8 & 59.7&-\\
    \midrule
    \multirow{10}*{\shortstack{22K Data\\ 26 GPU hours}}&\multicolumn{4}{l}{\emph{\textbf{Non-Pretrained Models}}}&&&\\
    &\small\texttt{3} & MM-GNN~\cite{gao2020multi} &Rosetta-en & 16.0 & - & 20.7 & 17.1\\
    &\small\texttt{4} & M4C~\cite{hu2020iterative} &Rosetta-en & 38.1 & 47.2 & 46.2 & 41.2\\
    &\small\texttt{5} & LaAP-Net~\cite{han2020finding} &Rosetta-en & 39.7 & 49.7 & 48.5 &-\\
    &\small\texttt{6} & SMA~\cite{gao2021structured} &Rosetta-en & - & - & 46.6 & -\\
    &\small\texttt{7} & SA-M4C~\cite{kant2020spatially} &Google-OCR & \textcolor{blue}{\textbf{42.2}} & \textcolor{blue}{\textbf{51.2}} & 49.5& \textcolor{blue}{\textbf{42.4}}\\
    &\small\texttt{8} & CRN~\cite{liu2020cascade} &Rosetta-en & - & - & 48.3&-\\
    &\small\texttt{9} & SSbaseline~\cite{zhu2020simple} &SBD-Trans & 40.0 & 50.8 & \textcolor{blue}{\textbf{49.7}} & 40.9\\
    &\small\texttt{10} & BOV~\cite{zeng2021beyond} &SBD-Trans & 40.2 & 50.0 & 47.2 &-\\
    &\small\texttt{11} & \textbf{DA-Net (Ours)} &SBD-Trans & \textcolor{red}{\textbf{48.5}} & \textcolor{red}{\textbf{57.5}} & \textcolor{red}{\textbf{53.0}}& \textcolor{red}{\textbf{50.0}}\\
    \bottomrule
    
    \end{tabular}
    \end{center}
\end{table*}

\vspace{-0.1in}
\subsection{Relation Prediction Task}
General models~\cite{zeng2021beyond,zhu2020simple} use linear layers to fuse the bounding box spatial information $B$ into visual features $f_v$.
\begin{gather}
    f^{obj} = LN(W_{v} f_{v}) + LN(W_{bx} B),
\end{gather}
where $W_{v}$ and $W_{bx}$ are trainable parameters.
However, the spatial fusion method lacks supervision and penalty. Thus, the model tends to ignore tokens' spatial information.

We design a relation prediction auxiliary task to add spatial supervision to object features to address these limitations. For object pairs in $Obj$, a relation prediction head generates an interrelation map $R^{N\times M}$ using bounding boxes and depths in $Obj$. As shown in Fig.~\ref{module_loss}, for $obj_i, obj_j \in Obj$, a relation prediction head uses a two-layer feed-forward network to generate their interrelation $r_{ij} \in Rel$.
\begin{gather}
    f^{obj} = LN(W_{v} f_{v}) + LN(W_{bx} [B^{obj}, d^{obj}])\\
    diff = W_{obj} f^{obj}_{j} - W_{obj} f^{obj}_{i},\\
    r_{ij} = W*diff + b,
\end{gather}
where $W_{obj}$ and $W$ are trainable parameters.

\textbf{Training Loss. }\quad To evaluate the accuracy of the inter-relation map $R$ between objects, we generate the ground-truth interrelation map $GT\in \mathbb{R}^{N\times M}$. Specifically, $gt_{ij} \in GT$ is $obj_i$ and $obj_j$'s ground-truth interrelation. Similar to $\Delta_{ij}$, $gt_{ij}$ is positively correlated with the cover rate between two object bounding boxes. $gt_{ij}$ is negatively correlated with the depth interpolation between objects.
\begin{gather}
    gt_{ij} = CR(B^{obj}_{i}, B^{obj}_{j}) *(d^{obj}_{j}-d^{obj}_{i}).
\end{gather}
We use mean multi-label Binary Cross-Entropy (BCE) loss to evaluate the similarity between the ground-truth relation matrix $gt$ and the predicted relation matrix $rel$:
\begin{gather}
    \mathcal{L}_{spatial} = BCE(gt, rel)
\end{gather}
To evaluate the accuracy of the answer, we follow the loss set from~\cite{zhu2020simple}. The semantic loss comes from two parts: the BCE loss and the Average Normalized Levenshtein Similarity (ANLS) loss: 
\begin{gather}
    \mathcal{L}_{semantic} = BCE(y_{pred}, y_{gt}) + ANLS(y_{pred}, y_{gt}),
\end{gather}
where $y_{pred}$ is the predicted answer and $y_{gt}$ is the ground truth answer. To this end, the total training loss is the weighted sum of $\mathcal{L}_{spatial}$ and $\mathcal{L}_{semantic}$.

The details of our DA-Net are shown in Algorithm~1. 
\begin{algorithm}[] 
	\caption{\textbf{DA-Net}}
	\LinesNumbered  
	\KwIn{Question embedding $q_i$, object and ocr features $f_i^{obj}, f_i^{ocr}$, object and ocr depth information $d^{obj}, d^{ocr}$, parameters $\Theta_{(i)}$ of DA-Net, the number of iterations $i=0$, learning rate $\lambda^{(i)}$ } 
	\While{Iter=i to total iteration}{
	    Compute the condensed question $Q_{obj}, Q_{ocr}$ by Equation~5,7\;
	    Compute the attention weight $S_{obj}, S_{ocr}$ by Equation~8,9\;
	    Compute the depth-aware weight transfer rate $\Delta$ by Equation~11\;
	    Compute the object and OCR's summary vectors $F^{obj}, F^{oct}$ by Equation~12,13\;
	    Fusing multimodal features $Q_{obj}, Q_{ocr}, F^{obj}, F^{oct}$ using the multi-modality transformer and predict the answer $y_{pred}$\;
	    Compute the loss $\mathcal{L}$ and gradients $x_i$ by Equation~20\;
	    Update parameters $\Theta_{(i+1)}$ by \:
            $\Theta_{(i+1)} = \Theta_{(i)} - \lambda^{(i)}\frac{\partial\mathcal{L}}{\partial x_i}\frac{\partial x_i}{\partial\Theta_{(i)}}$
	}
	\KwOut{parameters of DA-Net $\Theta_{(i+1)}$} 
\end{algorithm}

\vspace{-0.2in}
\section{Experiments}
We evaluate our model for the TextVQA task on two challenging datasets, including TextVQA~\cite{DBLP:conf/cvpr/SinghNSJCBPR19} and ST-VQA~\cite{BitenTMBRJVK19}. 
We also evaluate our model on the TextCaps dataset~\cite{sidorov2020textcaps} for text-based image captioning task. Experimental results show that our model achieves superior performance on both TextVQA and TextCaps tasks. All the performance of our model are public on the challenge website of \href{https://textvqa.org/challenge/}{TextVQA}, \href{https://rrc.cvc.uab.es/?ch=11}{STVQA} and \href{https://textvqa.org/textcaps/}{TextCaps}.

\vspace{-0.1in}
\subsection{Implementation Details}
Following M4C~\cite{hu2020iterative}, our input includes three parts: 20 question tokens, 100 object tokens, and 50 OCR tokens. For question features, we use three layers BERT~\cite{devlin2018bert} to extract features from question tokens. The BERT layers are finetuned during training. For object features, we use 
ResNet-152~\cite{xie2017aggregated} based Faster-RCNN model~\cite{ren2015faster} to extract object regions and their bounding boxes. For OCR features, we use Faster-RCNN model and SBD-Trans model~\cite{liu2019omnidirectional} to extract OCR visual features and OCR semantic features from images.

We use Adam as our optimizer.  The learning rate for TextVQA is set to $1e^{-4}$. We multiply the learning rate with a factor of $0.1$ at $14,000$ and $15,000$ iterations in a total of $24,000$ iterations. We set the maximum length of questions to $L = 20$. We recognize at most $M = 50$ OCR tokens and detect at most $N = 100$ objects. The maximum number of decoding steps is set to $12$. Transformer layer in our model uses $12$ attention heads. We use the same model on TextVQA and ST-VQA, only with different answer vocabulary, both with a fixed size of $5,000$. For the TextCaps task, following the M4C setting, we multiply the learning rate with a factor of $0.1$ at $3,000$ and $4,000$ iterations in a total of $12,000$ iterations. We set the maximum number of decoding steps to $30$.

\vspace{-0.1in}
\subsection{Comparison with The State-of-The-Art}

\textbf{TextVQA. } The TextVQA dataset~\cite{DBLP:conf/cvpr/SinghNSJCBPR19} contains $28,408$ images from the Open Images dataset~\cite{kuznetsova2020open}, with questions asking about text in the image. Each question in the TextVQA dataset has 10 free-response answers. Following the M4C~\cite{hu2020iterative}, we collect the top $5,000$ frequent words from the answers in the training set as answer vocabulary. 

Table~\ref{table:TextVQA} shows the performances of our model and other baselines. In the TextVQA task, the OCR recognition quality greatly affects the performance of the reasoning model. Following the previous state-of-the-art model BOV~\cite{zeng2021beyond}, we use the SBD-Trans OCR system as our OCR recognition backbone. Our DA-Net builds on the SSbaseline structure. Compared with SSbaseline, our model outperforms 2.4\% under TextVQA Val split and outperforms 1.9\% under Test split. Compared with the previous state-of-the-art model BOV, our model outperforms it by 1.5\% under TextVQA Val split and 1.0\% under Test split. We also use a classic OCR system, Rosetta-en, as our OCR backbone. As shown in line~14, our DA-Net also outperforms others under the Rosetta-en backbone.

Following previous TextVQA models~\cite{zeng2021beyond,gao2021structured,zhu2020simple}, we also train our model with additional ST-VQA training data. Our DA-Net also outperforms SA-M4C, SMA, SSbaseline and BOV, setting a new state-of-the-art result of \textbf{47.12}\% validation accuracy and \textbf{47.11}\% test accuracy on TextVQA dataset.

\begin{table}
    \vspace{-0.1in}
    \begin{tabular}{c|cc|cc}
    \toprule
    \multirow{2}{*}{Dataset} & \multicolumn{2}{c|}{\textbf{TextVQA}} & \multicolumn{2}{c}{\textbf{ST-VQA}}\\
    \cline{2-5} & Val & 3D Val & Val & 3D Val\\
    \midrule
    Image number & 5000 & 1262 & 2628 & 841\\
    Mean Question Length & 7.1 & 8.8 & 7.8 & 9.1\\
    Question Types & 14 & 12 & 12 & 8\\
    Vocabulary & 3450 & 1150 & 2300 & 868\\
    \midrule
    \end{tabular}
    \caption{Statistical summary of the 3D spatail subset.}
    \label{table:subset_analyse}
    \vspace{-0.1in}
\end{table}

\textbf{ST-VQA. } The ST-VQA dataset~\cite{BitenTMBRJVK19} is another recently proposed dataset for the TextVQA task. It contains 18,921 training and 2,971 test images sourced from COCO, ICDAR, IIIT, ImageNet, Visual Genome, and VizWiz datasets. Following M4C, we report results on the hardest Open Dictionary (Task-3) as it matches the TextVQA setting where no answer candidates are provided at test time.

Table~\ref{table:STVQA} shows the performance of our model and baselines on the ST-VQA dataset. We use SBD-trans as our OCR detection system.
Following prior works, we show our model's VQA accuracy and ANLS both on validation set and only the ANLS metric on the test set. On the validation set, our model achieves an accuracy of 48.5\% and an ANLS of 57.5\%,  which has 8.3\% and 7.5\% absolute higher than BOV. On the test set, our model achieves an ANLS of 53.0\%, 3.3\% better than SSbase and 5.8\% better than BOV. Overall, our model achieves the state-of-the-art performance on valid split and test split.

\textbf{Spatial Subsets of TextVQA and STVQA. }
This experiment is performed on spatial subsets to further explore the TextVQA models' ability to address spatial-related problems. Inspired by SAM4C~\cite{kant2020spatially}, we extract all the questions with ``on, top, under'' in the valid split of the TextVQA and STVQA datasets and generate the 3D subsets for both datasets. However, the SAM4C subset~\cite{kant2020spatially} considers spatial reasoning as well as OCR matching and contains 409 questions, while our 3D subset only considers 3D spatial reasoning and contains 1766 questions. The statistical summary of our 3D spatial subset is shown in Table~\ref{table:subset_analyse}. We re-evaluate our model and baselines on the new 3D subset. As shown in the ``Acc. on 3D Subset'' column of Table~\ref{table:TextVQA} and Table~\ref{table:STVQA}, DA-Net achieves the best results under both 3D subsets, surpassing the second by 3.8\% and 7.6\% on the TextVQA and STVQA subset.

\textbf{Comparison with Pretrained Models. }
As shown in Table I and Table II, our DA-Net (a non-pretrained method) achieves competitive performance compared to the pretrained models, for example, 48.5\% vs. 48.6\% by LOGOs [58] and 50.8\% by TAP [27] in accuracy, although these pretrained models spend quite large scale of training data and GPU hours. Specifically, compared to a non-pretrained method (i.e., our method), the scale of the training data for the pretrained models in the TextVQA task is about 70 times, and the GPU hours is about 20 times. In other words, our method is much more efficiency and cheaper compared to the pretrained models.

\begin{table*}[ht]
    \vspace{-0.1in}
    \caption{Ablation study for our depth-aware attention calibration module~(DAC) and relation prediction module~(Spatial Loss).}
    \label{table:ablation}
    \begin{tabular}{clcccc|cc}
    \toprule
     \# & Method & \makecell{Depth Info} &\makecell{Spatial Relation} & \makecell{DAC} & \makecell{Spatial Loss} & \makecell{Val Acc.} &\makecell{Acc. on\\Subset}\\
    \midrule
    \small\texttt{1}& SSbaseline & \XSolidBrush &2D & & & 43.9& 43.5 \\
    \small\texttt{2}&SSbaseline & \Checkmark &3D & & & 43.6{$_{\textcolor{blue}{(-0.3)}}$} & 43.7{$_{\textcolor{red}{(+0.2)}}$} \\
    \midrule
    \small\texttt{3}&Ablation & &2D+Overlap &  &  & 44.8{$_{\textcolor{red}{(+0.9)}}$}& 45.4{$_{\textcolor{red}{(+1.9)}}$} \\
    \small\texttt{4}&Ablation & &3D & Visual & & 45.1{$_{\textcolor{red}{(+1.2)}}$} & 47.6{$_{\textcolor{red}{(+4.1)}}$} \\
    \small\texttt{5}&Ablation & &3D & Semantic & & 44.7{$_{\textcolor{red}{(+0.8)}}$} &46.4{$_{\textcolor{red}{(+2.9)}}$} \\
    \small\texttt{6}&Ablation & &3D & Both & & 46.4{$_{\textcolor{red}{(+2.5)}}$} & 48.2{$_{\textcolor{red}{(+4.7)}}$} \\
    \small\texttt{7}&Ablation & &3D & & Bin Classification & 44.1{$_{\textcolor{red}{(+0.2)}}$} & 44.9{$_{\textcolor{red}{(+1.4)}}$} \\
    \small\texttt{8}&Ablation & &3D & & Regression & 45.0{$_{\textcolor{red}{(+1.1)}}$} &47.1{$_{\textcolor{red}{(+3.6)}}$} \\
    \midrule
    \small\texttt{9}&DA-Net & &3D & Full & Regression & \textbf{47.2}{$_{\textcolor{red}{(+3.3)}}$} & \textbf{49.2}{$_{\textcolor{red}{(+5.7)}}$} \\
    \bottomrule
    \end{tabular}
    \vspace{-0.1in}
\end{table*}

\begin{figure*}[ht]
    \vspace{-0.1in}
    \centering 
    \includegraphics[width=1\textwidth]{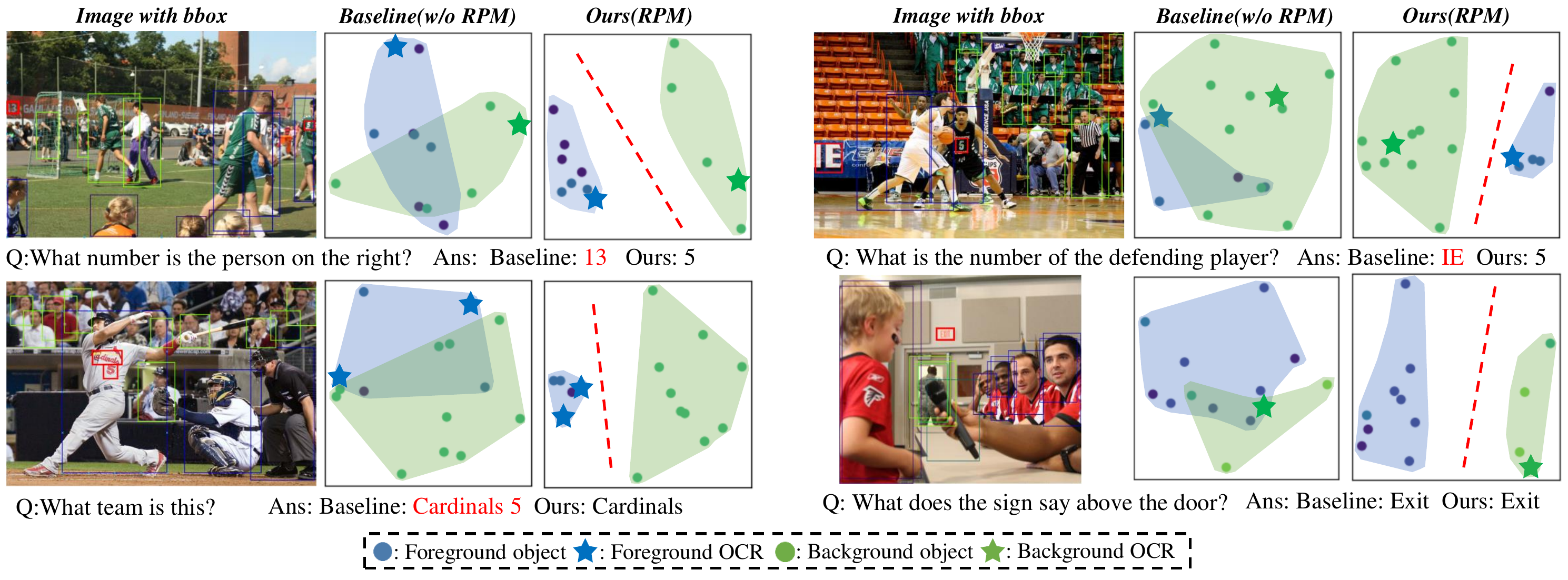} 
    \caption{Visualization of object's and OCR's spatial representations learned by the ssbaseline~(without relation prediction module~(RPM)) and our DA-Net~(with relation prediction module). Comparing the cluster map, our model can successfully split both foreground and background features while the baseline cannot. With RPM the model can learn a clear dividing line between foreground and background features. The RPM promotes the model's spatial understanding.} 
    \label{visual_3D} 
    \vspace{-0.1in}
\end{figure*}

\vspace{-0.1in}
\subsection{Ablation Study}
To further verify the performance of every module, we make ablation studies on TextVQA valid split and its 3D subset. Table~\ref{table:ablation} shows every module's influence on the total model. Fig~\ref{statistical_fig} shows the model's performance under various degrees of noise interference.

\textbf{Effect of Depth Information. } To verify the direct improvement of model performance by adding depth information, we remove our depth-aware attention calibration module and the relation prediction task, only using depth information to supplement the 2D bounding box. As shown in Table~\ref{table:ablation} line~2, our baseline model achieves 43.6\% on the validation split. Compared with the baseline, our model drops 0.3\% because the model cannot comprehend the meaning of depth information without supervision. When bringing the depth information directly into the model, the depth information may be treated as noise information, and thus disturbs the model's spatial reasoning process. 

\textbf{Effect of Depth-aware Attention Calibration Module. } To verify the effectiveness of our depth-aware attention calibration (DAC) module, we remove the relation prediction task and apply the depth-aware attention calibration module on the OCR-visual part~(Visual), OCR-semantic part~(Semantic), and on both parts~(Both).
Our model achieves 45.1\% when adding depth-aware attention calibration on the OCR-visual part and achieves 44.7\% when adding calibration on the OCR-semantic part. Compared with the baseline, our model has a 0.8-1.2\% improvement on validation split. When adding the attention calibration module on both OCR parts, our full depth-aware attention calibration module achieves 46.4\% with a 2.5\% improvement. 
The performance of the DAC module is more obvious on the 3D subset of the validation split. This proves the accuracy increase mainly comes from calibrating the OCR attention score according to the spatial relation of the critical object in spatial-related questions.
As analyzed in the following subsection, our depth-aware attention calibration module can transfer the attention weight from key objects to their spatial corresponding text tokens. The DAC module guides the model toward the potential answers and restrains from distractor tokens.

\begin{figure*}[ht] 
    \vspace{-0.1in}
    \centering 
    \includegraphics[width=1\textwidth]{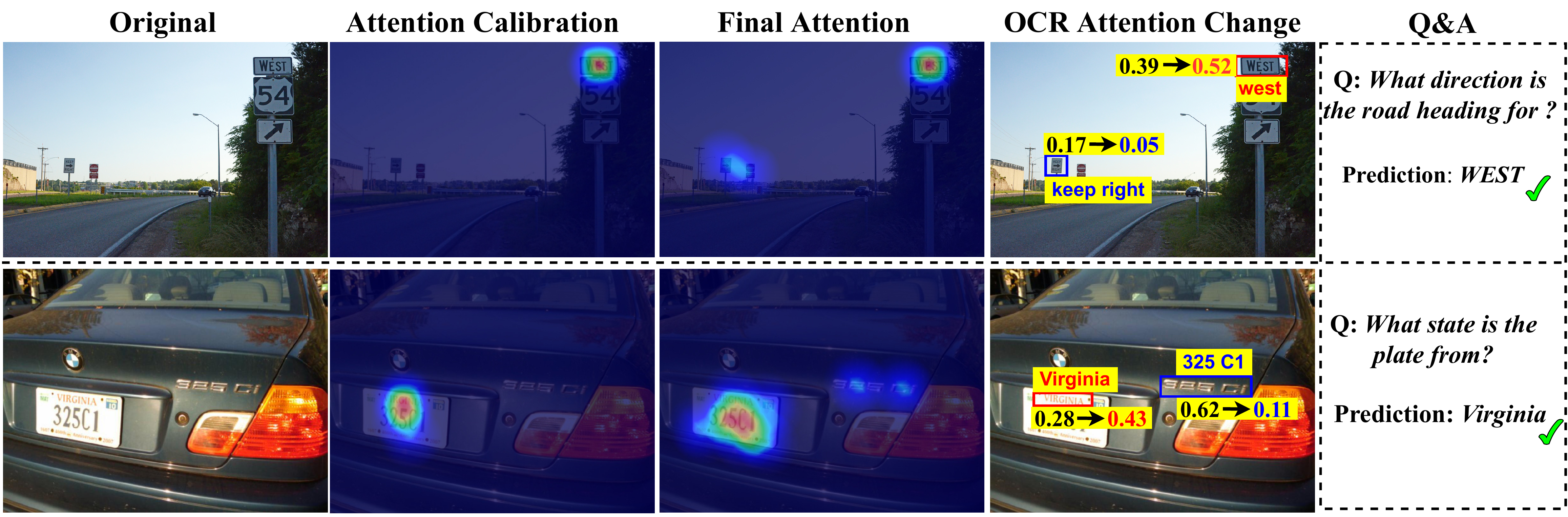} 
    \vspace{-0.1in}
    \caption{Visualization of OCR attention scores for our model with and without the Depth-aware Attention Calibration~(DAC). \textit{Attention Calibration} represents the visualization of additional scores after the calibration of our module. Our DAC successfully calibrates our model's OCR attention and focuses on the correct OCR token in the two cases.
    \textit{Final Attention} is the visualization of all OCR's attention scores after the calibration of our module.
    \textit{OCR Attention Change} shows the OCR attention score change with and without the depth-aware attention calibration. Our DAC can make the proper calibration, improving the score of correct OCR~(\textcolor{red}{Red}) and reducing the distractor OCR score~(\textcolor{blue}{Blue}).
    } 
    \label{visual_summary} 
    \vspace{-0.1in}
\end{figure*}


\textbf{Effect of Relation Prediction Module. } To verify the effectiveness of our relation prediction auxiliary loss, we remove the depth-aware attention calibration module from the whole model. As shown in Table~\ref{table:ablation}, our model with only relation prediction modification achieves 45.0\% ANLS on validation split, outperforming the baseline by 1.1\%. In the 3D subset validation, our model with relation prediction modification surpasses the baseline at 3.6\%.  The results shows that adding the relation prediction task helps the model to understand the 3D geometric information. Inspired by Adabins~\cite{bhat2021adabins}, we also change the spatial loss into bin classification form. Results show that bin classification is worse than the regression form.

\textbf{Robustness to Depth Noise. }
The depth information predicted by AdaBins~\cite{bhat2021adabins} contains slight noise. To verify the depth noise influence on our model, we randomly mask the input depth information and Table~\ref{statistical_fig} shows the results. Specifically, we replace the original depth information using the mean value of all depth information as the noise. After masking 80\% and 40\% depth information, the performance drops significantly. However, after masking 20\% and 10\% depth information, the performance only suffers little volatility, outperforms the SSbaseline model. It indicates that our model is able to suffer slight depth noise.

\vspace{-0.1in}
\subsection{Qualitative Analysis}
In this part, we cluster the object and OCR features in high dimensions to show the model's 3D understanding improvement after the addition of the relation prediction task. We visualize heat maps of objects and OCR attention scores to demonstrate the effectiveness of our depth-aware attention calibration module. And we select representative cases to show our model's ability to address spatial-related questions. We use the baseline model: SSbaseline for comparison.

\textbf{Relation Prediction Module: } 
To demonstrate that our relation prediction module can strengthen the model's understanding of the 3D spatial relationship between objects, we visualize the spatial clustering of objects and OCRs with and without the relation prediction module in Fig.~\ref{visual_3D}. 
We choose images with obvious foreground and background split as our visualization cases. \textcolor[rgb]{0.2,0.65,0}{\textbf{Green}} represents objects and OCRs in background and \textcolor{blue}{\textbf{blue}} represents objects and OCRs in foreground. To ensure that the clustering is only related to spatial information instead of semantic information, we only visualize objects with the same class: person. We use PCA as the visualization tool.

As shown in Fig.~\ref{visual_3D}, without the relation prediction module, the baseline cannot distinguish the spatial relation between foreground and background~(the overlap between green and blue area). After adding the relation prediction module, our model successfully splits the features from the foreground and background. The clustering shows the relation prediction module helps the reasoning model to establish an implicit spatial structure. With RPN, our model has a better 3D understanding of the TextVQA task.

\textbf{Depth-aware Attention Calibration (DAC): } Our depth-aware attention calibration module transfers the object's attention score to calibrate OCR's attention score under the supervision of 3D spatial relations. This procedure enables the model to calibrate the OCR's attention using critical object information.
In \textit{Attention Calibration}, the second column of Fig.~\ref{visual_summary}, we visualize the OCR's additional attention scores during the transferring process. To make the picture clearer, we show the heatmap of the top three additional attention scores.

\begin{figure*}[ht] 
    \centering 
    \includegraphics[width=1\textwidth]{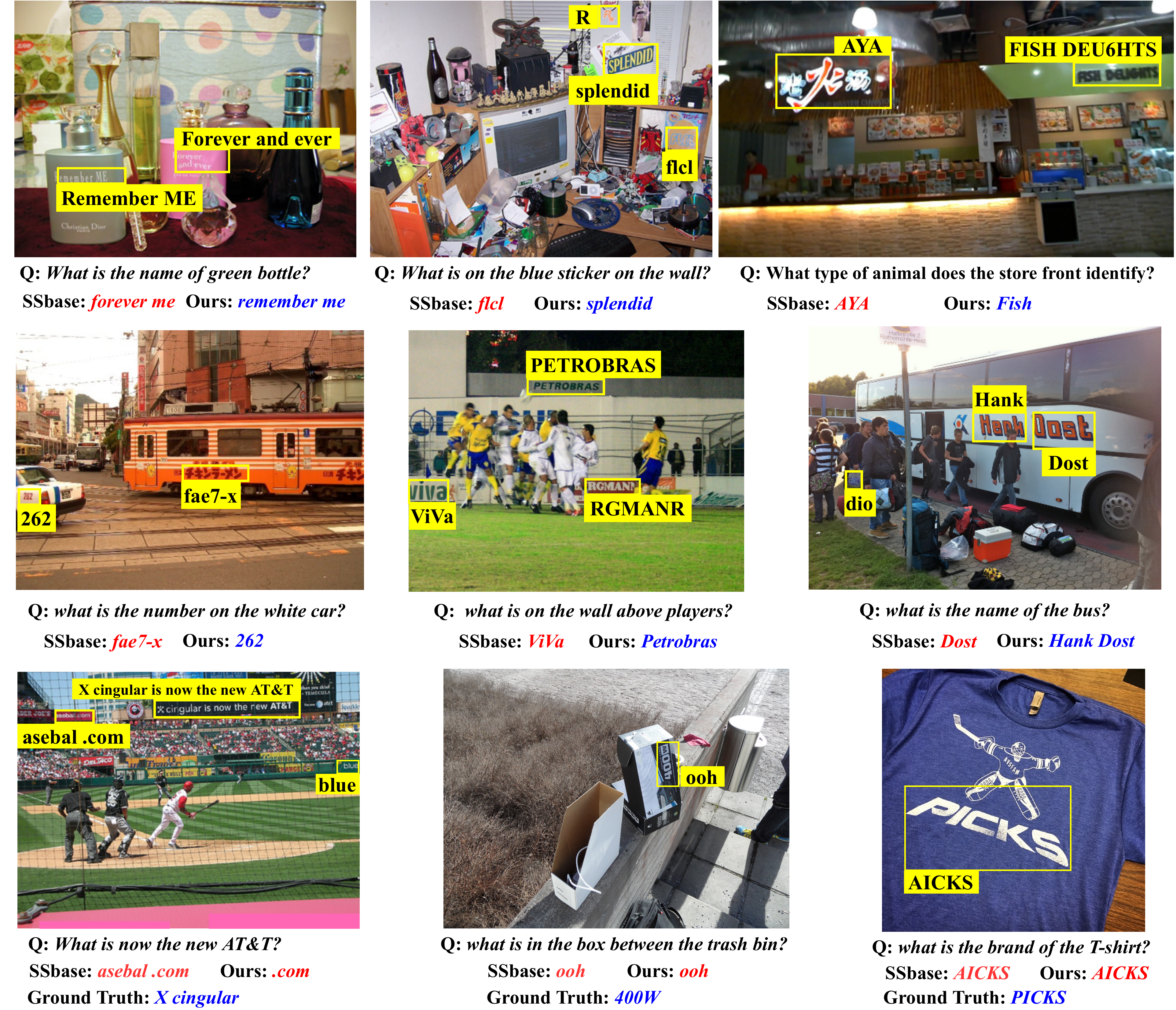} 
    \vspace{-0.1in}
    \caption{The visualization for example results generated by our model~(Ours) and the baseline model~(SSbase). With the supervision of depth information and our step-by-step 3D spatial reasoning procedure, our model can better predict answers involving object-text spatial relations and better locate correct objects. The \textcolor{red}{Red} text denote the wrong answers and the \textcolor{blue}{Blue} denote the right answers. However, both our model and the baseline fail to deal with cases with incorrect OCR tokens. } 
    \label{visual_badcase} 
    \vspace{-0.1in}
\end{figure*}

To show the change of attention scores after adding our DAC module, we visualize the OCR's attention scores before and after transferring process in \textit{OCR Attention Change}, the fourth column of Fig.~\ref{visual_summary}. We only show the top two OCR tokens with noticeable attention score changes to make the picture clearer.
For both two cases, the transferred attention score correctly locates the key OCR tokens~(``west'' and ``virginia''). As the \textit{OCR Attention Change} column shows, for the first case, the attention score of the correct answer token ``West'' increases by 0.13 after adding the DAC module. For the second case, we find out that our DAC module can also restrain the attention score for distractor token~(``325 C1''), which guides the model to answer the correct answer.


\begin{table*}[htbp]
\resizebox{0.93\textwidth}{!}{
    \begin{floatrow}
    \capbtabbox{
     \begin{tabular}{lcccc}
        \toprule
        Method & \makecell{Layer} & \makecell{Attention Block} & \makecell{Transformer Encoder} & \makecell{Total in Encoder~(FLOPs)} \\
        \midrule
         M4C & per-layer & - & $(L+N+M)^2$ & 28,900 \\
         M4C & 4-layer & - & $4(L+N+M)^2$ & 115,600 \\
         SSbase & per-layer & $2(2N+M)$ & $(6+N)^2$ & 3,136 \\
         SSbase & 8-layer & $2(2N+M)$ & $8(6+N)^2$ & 25,488 \\
         BOV & per-layer & $N+M$ & $(L+N+M+C)^2$ & 32,550\\
         BOV & 4-layer & $N+M$ & $4(L+N+M+C)^2$ & 129,750\\
         \midrule
         DA-Net & per-layer & $2(2N+2M)$ & $(6+N)^2$ & 3,336\\
         DA-Net & 8-layer & $2(2N+2M)$ & $8(6+N)^2$ & 25,288\\
        \bottomrule
    \end{tabular}
    }{
     \vspace{-0.1in}
     \caption{Computational complexity~(FLOPs) of two model encoders.}
     \label{table:computation}
    }
    \capbtabbox{
    \begin{tabular}{ll}
      \toprule
      \textbf{Noise}&\textbf{Acc.}\\
      \midrule
      \textbf{SSbaseline} & \\
       ~\emph{No Noise}& 43.9 \\
      \midrule
      \textbf{DA-Net} & \\
       ~\emph{No Noise} & 47.2{$_{\textcolor{red}{(+3.3)}}$} \\
       ~\emph{10\% Noise} & 45.5{$_{\textcolor{red}{(+1.6)}}$}\\
       ~\emph{20\% Noise} & 45.3{$_{\textcolor{red}{(+1.4)}}$}\\
       ~\emph{40\% Noise} & 43.1{$_{\textcolor{blue}{(-0.8)}}$}\\
       ~\emph{80\% Noise} & 42.2{$_{\textcolor{blue}{(-1.7)}}$}\\
      \bottomrule
    \end{tabular}
    }{
     \vspace{-0.1in}
     \caption{DA-Net robustness to noise.}
     \label{statistical_fig}
    }
    \end{floatrow}
}
\end{table*}

\begin{table}
\resizebox{0.98\textwidth}{!}{
    \caption{The performance on TextCaps~\cite{sidorov2020textcaps}.}
    \vspace{-0.1in}
    \label{table:TextCaps}
    \begin{tabular}{lccccc}
    
    \toprule
    Method & \makecell{BLEU-4} & \makecell{METEOR} & \makecell{ROUGE} & \makecell{SPICE} & \makecell{CIDEr}\\
    \midrule
     CNMT\cite{wang2021confidence} & 19.97 & 20.91 & 44.37 & 13.52 &  \textcolor{blue}{\textbf{93.03}}\\
     BUTD\cite{anderson2018bottom} &  20.10 & 17.80 & 42.90 & 11.70 & 41.90\\
     UIT\cite{UIT} & 20.02 & 20.89 & 44.41 & 13.74 & 85.64\\
     AoANet\cite{huang2019attention} & 20.40 & 18.90 & 42.90 & 13.20 & 42.70\\
     M4C-Cap\cite{hu2020iterative} & 23.30 & 22.00 & 46.20 & \textcolor{red}{\textbf{15.53}} & 89.60\\
     SSbase\cite{zhu2020simple} & \textcolor{blue}{\textbf{23.53}} & \textcolor{blue}{\textbf{22.02}} & \textcolor{blue}{\textbf{46.40}} & 15.01 & 90.50\\
    \midrule
     DA-Net & \textcolor{red}{\textbf{24.35}} & \textcolor{red}{\textbf{22.47}} & \textcolor{red}{\textbf{46.95}} & \textcolor{blue}{\textbf{15.32}} & \textcolor{red}{\textbf{96.80}}\\
    \bottomrule
    \end{tabular}
}
\end{table}

\textbf{Case Study. }  About 44.1\% of the questions in the TextVQA dataset need one or more spatial reasoning~(Fig.~\ref{figure3}). We present several cases with questions that need spatial reasoning for further analysis. As shown in Fig.~\ref{visual_badcase}, our model performs well in challenging cases with the supervision of 3D spatial information and a step-by-step spatial reasoning procedure. 

The first case on line~1 and line~2 of Fig.~\ref{visual_badcase} shows that our model can better predict OCR answers with object-text spatial relations, while the baseline fails to generate the entirely correct answer. 
The second and third cases on line~1 and line~2 of Fig.~\ref{visual_badcase} show that our model can better locate correct objects while the baseline tends to locate the distractors.

However, there are cases in which both our DA-Net and the baseline cannot predict the answer correctly. (i) Images with complex semantic information in OCR tokens. As shown in the first case on line~3, when questions aim at the semantic information in the OCR token, existing TextVQA models tend to predict irrelated OCR in the image as the answer. (ii) Images with false OCR information. As shown in the second and third cases on line~3, when the OCR detection module generates incorrect OCR tokens, our downstream reasoning model cannot predict correct answers.

\vspace{-0.1in}
\subsection{Computational Complexity} 
We list the computational complexity of our DA-Net and other TextVQA models including SSbaseline, M4C, and BOV in Table~\ref{table:computation}. Specifically, $L=20$ is the length of the question; $N=50$ is the number of the OCR tokens; $M=100$ is the number of the detected objects; $C=10$ is the number of the candidate answers in BOV. We omit all vector dimensions D for simplicity. 

Due to the addition of the depth-aware attention calibration module, our DA-Net's attention block complexity is slightly larger than other models. Table~\ref{table:computation} also shows the complexity of the MMT encoder and the comparison of the encoder's total computation complexity~(the summary of Transformer encoder and attention block complexity). The computational complexity of our DA-Net structure is much significantly smaller than the complexity of M4C and BOV. Our depth-aware attention calibration module and the relation prediction module create a little computation complexity increase.

\vspace{-0.2in}
\textbf{\subsection{Generalization Analysis}}
In addition to TextVQA, our model can also be applied to other scene-text tasks including the TextCaps task. This task requires models to generate image descriptions via texts in the context of images. In this paper, we conduct experiments on the TextCaps dataset~\cite{sidorov2020textcaps} for text-based image captioning. The TextCaps dataset, with 145k captions for 28k images, is recently proposed for the text-based image caption task. 

Following M4C-Captioner~\cite{sidorov2020textcaps}, we use BLEU-4~\cite{papineni2002bleu}, METEOR~\cite{denkowski2014meteor}, ROUGE~\cite{rouge2004package}, SPICE~\cite{anderson2016spice} and CIDEr~\cite{vedantam2015cider} to evaluate the performance of captioning models. All automatic metrics are positively correlated with the generated quality. Table~\ref{table:TextCaps} shows that DA-Net outperforms others and achieves a new state-of-the-art performance of 96.80 CIDEr-D score and 24.35 BLEU-4 score.
The performance demonstrates our model's generalization ability on other scene-text tasks. 

\vspace{+0.2in}
\section{Conclusion and Future Work}
\label{sec:conclusion}
In this paper, we study the spatial reasoning problem from human-like spatial reasoning for TextVQA. To this end, we propose the Depth-Aware TextVQA Network (DA-Net) by introducing the 3D depth information into the TextVQA task. We introduce a relation prediction module to enhance the model's understanding of 3D spatial relations. Besides, we design a depth-aware attention calibration module to readjust the distribution of OCR tokens' attention scores based on the crucial object. The experiments on two TextVQA datasets demonstrate the effectiveness of our method. Further, we also apply 3D spatial reasoning in our work to other visual reasoning tasks such as text-based image captioning.

\textbf{Limitation}: The main limitation of DA-Net is that the depth estimation model could produce noisy depth information and is not well designed for TextVQA. Our future work will integrate the depth estimation module into the DA-Net to adapt the TextVQA task in an end-to-end manner, which may further boost the spatial reasoning ability of our model.

\clearpage

\bibliographystyle{ieeetr}
\bibliography{egbib}

\begin{thebibliography}{10}

\bibitem{mishra2019ocr}
A.~Mishra, S.~Shekhar, A.~K. Singh, and A.~Chakraborty, ``Ocr-vqa: Visual
  question answering by reading text in images,'' in {\em 2019 International
  Conference on Document Analysis and Recognition (ICDAR)}, pp.~947--952, IEEE,
  2019.

\bibitem{BitenTMBRJVK19}
A.~F. Biten, R.~Tito, A.~Mafla, L.~G. i~Bigorda, M.~Rusi{\~{n}}ol, C.~V.
  Jawahar, E.~Valveny, and D.~Karatzas, ``Scene text visual question
  answering,'' 2019.
\newblock 2019 {IEEE/CVF} International Conference on Computer Vision, {ICCV}
  2019, Seoul, Korea (South), October 27 - November 2, 2019.

\bibitem{singh2019towards}
A.~Singh, V.~Natarajan, M.~Shah, Y.~Jiang, X.~Chen, D.~Batra, D.~Parikh, and
  M.~Rohrbach, ``Towards vqa models that can read,'' in {\em Proceedings of the
  IEEE/CVF Conference on Computer Vision and Pattern Recognition},
  pp.~8317--8326, 2019.

\bibitem{qiu2019incorporating}
Y.~Qiu, Y.~Satoh, R.~Suzuki, and H.~Kataoka, ``Incorporating 3d information
  into visual question answering,'' in {\em 2019 International Conference on 3D
  Vision (3DV)}, pp.~756--765, IEEE, 2019.

\bibitem{zeng2021beyond}
G.~Zeng, Y.~Zhang, Y.~Zhou, and X.~Yang, ``Beyond ocr+ vqa: Involving ocr into
  the flow for robust and accurate textvqa,'' in {\em Proceedings of the 29th
  ACM International Conference on Multimedia}, pp.~376--385, 2021.

\bibitem{zhu2020simple}
Q.~Zhu, C.~Gao, P.~Wang, and Q.~Wu, ``Simple is not easy: A simple strong
  baseline for textvqa and textcaps,'' {\em arXiv preprint arXiv:2012.05153},
  vol.~2, 2020.

\bibitem{gao2021structured}
C.~Gao, Q.~Zhu, P.~Wang, H.~Li, Y.~Liu, A.~Van~den Hengel, and Q.~Wu,
  ``Structured multimodal attentions for textvqa,'' {\em IEEE Transactions on
  Pattern Analysis and Machine Intelligence}, 2021.

\bibitem{kant2020spatially}
Y.~Kant, D.~Batra, P.~Anderson, A.~Schwing, D.~Parikh, J.~Lu, and H.~Agrawal,
  ``Spatially aware multimodal transformers for textvqa,'' in {\em European
  Conference on Computer Vision}, pp.~715--732, Springer, 2020.

\bibitem{helwe2021reasoning}
C.~Helwe, C.~Clavel, and F.~M. Suchanek, ``Reasoning with transformer-based
  models: Deep learning, but shallow reasoning,'' in {\em 3rd Conference on
  Automated Knowledge Base Construction}, 2021.

\bibitem{krishna2017visual}
R.~Krishna, Y.~Zhu, O.~Groth, J.~Johnson, K.~Hata, J.~Kravitz, S.~Chen,
  Y.~Kalantidis, L.-J. Li, D.~A. Shamma, {\em et~al.}, ``Visual genome:
  Connecting language and vision using crowdsourced dense image annotations,''
  {\em International journal of computer vision}, vol.~123, no.~1, pp.~32--73,
  2017.

\bibitem{DBLP:conf/cvpr/SinghNSJCBPR19}
A.~Singh, V.~Natarajan, M.~Shah, Y.~Jiang, X.~Chen, D.~Batra, D.~Parikh, and
  M.~Rohrbach, ``Towards {VQA} models that can read,'' in {\em {IEEE}
  Conference on Computer Vision and Pattern Recognition, {CVPR} 2019, Long
  Beach, CA, USA, June 16-20, 2019}, pp.~8317--8326, 2019.

\bibitem{banerjee2021weakly}
P.~Banerjee, T.~Gokhale, Y.~Yang, and C.~Baral, ``Weakly supervised relative
  spatial reasoning for visual question answering,'' in {\em Proceedings of the
  IEEE/CVF International Conference on Computer Vision}, pp.~1908--1918, 2021.

\bibitem{vieu1997spatial}
L.~Vieu, ``Spatial representation and reasoning in artificial intelligence,''
  in {\em Spatial and temporal reasoning}, pp.~5--41, Springer, 1997.

\bibitem{chang2014learning}
A.~Chang, M.~Savva, and C.~D. Manning, ``Learning spatial knowledge for text to
  3d scene generation,'' in {\em Proceedings of the 2014 conference on
  empirical methods in natural language processing (EMNLP)}, pp.~2028--2038,
  2014.

\bibitem{zhou2022sp}
Y.~Zhou, W.~Xiang, C.~Li, B.~Wang, X.~Wei, L.~Zhang, M.~Keuper, and X.~Hua,
  ``Sp-vit: Learning 2d spatial priors for vision transformers,'' {\em arXiv
  preprint arXiv:2206.07662}, 2022.

\bibitem{wu2021cvt}
H.~Wu, B.~Xiao, N.~Codella, M.~Liu, X.~Dai, L.~Yuan, and L.~Zhang, ``Cvt:
  Introducing convolutions to vision transformers,'' in {\em Proceedings of the
  IEEE/CVF International Conference on Computer Vision}, pp.~22--31, 2021.

\bibitem{yuan2021incorporating}
K.~Yuan, S.~Guo, Z.~Liu, A.~Zhou, F.~Yu, and W.~Wu, ``Incorporating convolution
  designs into visual transformers,'' in {\em Proceedings of the IEEE/CVF
  International Conference on Computer Vision}, pp.~579--588, 2021.

\bibitem{yuan2021tokens}
L.~Yuan, Y.~Chen, T.~Wang, W.~Yu, Y.~Shi, Z.-H. Jiang, F.~E. Tay, J.~Feng, and
  S.~Yan, ``Tokens-to-token vit: Training vision transformers from scratch on
  imagenet,'' in {\em Proceedings of the IEEE/CVF International Conference on
  Computer Vision}, pp.~558--567, 2021.

\bibitem{li2022joint}
H.~Li, X.~Li, B.~Karimi, J.~Chen, and M.~Sun, ``Joint learning of object graph
  and relation graph for visual question answering,'' {\em arXiv preprint
  arXiv:2205.04188}, 2022.

\bibitem{johnson2017clevr}
J.~Johnson, B.~Hariharan, L.~Van Der~Maaten, L.~Fei-Fei, C.~Lawrence~Zitnick,
  and R.~Girshick, ``Clevr: A diagnostic dataset for compositional language and
  elementary visual reasoning,'' in {\em Proceedings of the IEEE conference on
  computer vision and pattern recognition}, pp.~2901--2910, 2017.

\bibitem{cao2019interpretable}
Q.~Cao, X.~Liang, B.~Li, and L.~Lin, ``Interpretable visual question answering
  by reasoning on dependency trees,'' {\em IEEE transactions on pattern
  analysis and machine intelligence}, vol.~43, no.~3, pp.~887--901, 2019.

\bibitem{jiang2018pythia}
Y.~Jiang, V.~Natarajan, X.~Chen, M.~Rohrbach, D.~Batra, and D.~Parikh, ``Pythia
  v0. 1: the winning entry to the vqa challenge 2018,'' {\em arXiv preprint
  arXiv:1807.09956}, 2018.

\bibitem{hu2020iterative}
R.~Hu, A.~Singh, T.~Darrell, and M.~Rohrbach, ``Iterative answer prediction
  with pointer-augmented multimodal transformers for textvqa,'' in {\em
  Proceedings of the IEEE/CVF Conference on Computer Vision and Pattern
  Recognition}, pp.~9992--10002, 2020.

\bibitem{chen2020uniter}
Y.-C. Chen, L.~Li, L.~Yu, A.~El~Kholy, F.~Ahmed, Z.~Gan, Y.~Cheng, and J.~Liu,
  ``Uniter: Universal image-text representation learning,'' in {\em European
  conference on computer vision}, pp.~104--120, Springer, 2020.

\bibitem{li2020oscar}
X.~Li, X.~Yin, C.~Li, P.~Zhang, X.~Hu, L.~Zhang, L.~Wang, H.~Hu, L.~Dong,
  F.~Wei, {\em et~al.}, ``Oscar: Object-semantics aligned pre-training for
  vision-language tasks,'' in {\em European Conference on Computer Vision},
  pp.~121--137, Springer, 2020.

\bibitem{lu2019vilbert}
J.~Lu, D.~Batra, D.~Parikh, and S.~Lee, ``Vilbert: Pretraining task-agnostic
  visiolinguistic representations for vision-and-language tasks,'' {\em
  Advances in neural information processing systems}, vol.~32, 2019.

\bibitem{yang2021tap}
Z.~Yang, Y.~Lu, J.~Wang, X.~Yin, D.~Florencio, L.~Wang, C.~Zhang, L.~Zhang, and
  J.~Luo, ``Tap: Text-aware pre-training for textvqa and text-caption,'' in
  {\em Proceedings of the IEEE/CVF Conference on Computer Vision and Pattern
  Recognition}, pp.~8751--8761, 2021.

\bibitem{wang2021improving}
J.~Wang, C.~Yang, Y.~Xu, Y.~Shen, H.~Li, and B.~Zhou, ``Improving gan
  equilibrium by raising spatial awareness,'' {\em arXiv preprint
  arXiv:2112.00718}, 2021.

\bibitem{liu2021learning}
N.~Liu, S.~Li, Y.~Du, J.~Tenenbaum, and A.~Torralba, ``Learning to compose
  visual relations,'' {\em Advances in Neural Information Processing Systems},
  vol.~34, 2021.

\bibitem{wu2017visual}
Q.~Wu, D.~Teney, P.~Wang, C.~Shen, A.~Dick, and A.~van~den Hengel, ``Visual
  question answering: A survey of methods and datasets,'' {\em Computer Vision
  and Image Understanding}, vol.~163, pp.~21--40, 2017.

\bibitem{sun2021video}
G.~Sun, L.~Liang, T.~Li, B.~Yu, M.~Wu, and B.~Zhang, ``Video question
  answering: a survey of models and datasets,'' {\em Mobile Networks and
  Applications}, vol.~26, no.~5, pp.~1904--1937, 2021.

\bibitem{li2019relation}
L.~Li, Z.~Gan, Y.~Cheng, and J.~Liu, ``Relation-aware graph attention network
  for visual question answering,'' in {\em Proceedings of the IEEE/CVF
  international conference on computer vision}, pp.~10313--10322, 2019.

\bibitem{yao2018exploring}
T.~Yao, Y.~Pan, Y.~Li, and T.~Mei, ``Exploring visual relationship for image
  captioning,'' in {\em Proceedings of the European conference on computer
  vision (ECCV)}, pp.~684--699, 2018.

\bibitem{narasimhan2018out}
M.~Narasimhan, S.~Lazebnik, and A.~Schwing, ``Out of the box: Reasoning with
  graph convolution nets for factual visual question answering,'' {\em Advances
  in neural information processing systems}, vol.~31, 2018.

\bibitem{ryoo2021tokenlearner}
M.~S. Ryoo, A.~Piergiovanni, A.~Arnab, M.~Dehghani, and A.~Angelova,
  ``Tokenlearner: What can 8 learned tokens do for images and videos?,'' {\em
  arXiv preprint arXiv:2106.11297}, 2021.

\bibitem{yang2020trrnet}
X.~Yang, G.~Lin, F.~Lv, and F.~Liu, ``Trrnet: Tiered relation reasoning for
  compositional visual question answering,'' in {\em European Conference on
  Computer Vision}, pp.~414--430, Springer, 2020.

\bibitem{gao2020multi}
D.~Gao, K.~Li, R.~Wang, S.~Shan, and X.~Chen, ``Multi-modal graph neural
  network for joint reasoning on vision and scene text,'' in {\em Proceedings
  of the IEEE/CVF Conference on Computer Vision and Pattern Recognition},
  pp.~12746--12756, 2020.

\bibitem{saxena2005learning}
A.~Saxena, S.~Chung, and A.~Ng, ``Learning depth from single monocular
  images,'' {\em Advances in neural information processing systems}, vol.~18,
  2005.

\bibitem{eigen2014depth}
D.~Eigen, C.~Puhrsch, and R.~Fergus, ``Depth map prediction from a single image
  using a multi-scale deep network,'' {\em Advances in neural information
  processing systems}, vol.~27, 2014.

\bibitem{li2017two}
J.~Li, R.~Klein, and A.~Yao, ``A two-streamed network for estimating
  fine-scaled depth maps from single rgb images,'' in {\em Proceedings of the
  IEEE International Conference on Computer Vision}, pp.~3372--3380, 2017.

\bibitem{bhat2021adabins}
S.~F. Bhat, I.~Alhashim, and P.~Wonka, ``Adabins: Depth estimation using
  adaptive bins,'' pp.~4009--4018, 2021.

\bibitem{hong2022depth}
F.-T. Hong, L.~Zhang, L.~Shen, and D.~Xu, ``Depth-aware generative adversarial
  network for talking head video generation,'' {\em arXiv preprint
  arXiv:2203.06605}, 2022.

\bibitem{shi20223d}
Z.~Shi, Y.~Shen, J.~Zhu, D.-Y. Yeung, and Q.~Chen, ``3d-aware indoor scene
  synthesis with depth priors,'' {\em arXiv preprint arXiv:2202.08553}, 2022.

\bibitem{ma2019accurate}
X.~Ma, Z.~Wang, H.~Li, P.~Zhang, W.~Ouyang, and X.~Fan, ``Accurate monocular 3d
  object detection via color-embedded 3d reconstruction for autonomous
  driving,'' in {\em Proceedings of the IEEE/CVF International Conference on
  Computer Vision}, pp.~6851--6860, 2019.

\bibitem{wang2019pseudo}
Y.~Wang, W.-L. Chao, D.~Garg, B.~Hariharan, M.~Campbell, and K.~Q. Weinberger,
  ``Pseudo-lidar from visual depth estimation: Bridging the gap in 3d object
  detection for autonomous driving,'' in {\em Proceedings of the IEEE/CVF
  Conference on Computer Vision and Pattern Recognition}, pp.~8445--8453, 2019.

\bibitem{weng2019monocular}
X.~Weng and K.~Kitani, ``Monocular 3d object detection with pseudo-lidar point
  cloud,'' in {\em Proceedings of the IEEE/CVF International Conference on
  Computer Vision Workshops}, pp.~0--0, 2019.

\bibitem{li2022locality}
K.~Li, R.~Yu, Z.~Wang, L.~Yuan, G.~Song, and J.~Chen, ``Locality guidance for
  improving vision transformers on tiny datasets,'' {\em arXiv preprint
  arXiv:2207.10026}, 2022.

\bibitem{jin2023diffusionret}
P.~Jin, H.~Li, Z.~Cheng, K.~Li, X.~Ji, C.~Liu, L.~Yuan, and J.~Chen,
  ``Diffusionret: Generative text-video retrieval with diffusion model,'' {\em
  arXiv preprint arXiv:2303.09867}, 2023.

\bibitem{liu2021refer}
H.~Liu, A.~Lin, X.~Han, L.~Yang, Y.~Yu, and S.~Cui, ``Refer-it-in-rgbd: A
  bottom-up approach for 3d visual grounding in rgbd images,'' in {\em
  Proceedings of the IEEE/CVF Conference on Computer Vision and Pattern
  Recognition}, pp.~6032--6041, 2021.

\bibitem{cheng2023parallel}
Z.~Cheng, K.~Li, P.~Jin, X.~Ji, L.~Yuan, C.~Liu, and J.~Chen, ``Parallel vertex
  diffusion for unified visual grounding,'' {\em arXiv preprint
  arXiv:2303.07216}, 2023.

\bibitem{ye20213d}
S.~Ye, D.~Chen, S.~Han, and J.~Liao, ``3d question answering,'' {\em arXiv
  preprint arXiv:2112.08359}, 2021.

\bibitem{li2023tg}
H.~Li, P.~Jin, Z.~Cheng, S.~Zhang, K.~Chen, Z.~Wang, C.~Liu, and J.~Chen,
  ``Tg-vqa: Ternary game of video question answering,'' {\em arXiv preprint
  arXiv:2305.10049}, 2023.

\bibitem{zhou20183d}
S.~Zhou and S.~Xiao, ``3d face recognition: a survey,'' {\em Human-centric
  Computing and Information Sciences}, vol.~8, no.~1, pp.~1--27, 2018.

\bibitem{han2020finding}
W.~Han, H.~Huang, and T.~Han, ``Finding the evidence: Localization-aware answer
  prediction for text visual question answering,'' {\em arXiv preprint
  arXiv:2010.02582}, 2020.

\bibitem{wang2022tag}
J.~Wang, M.~Gao, Y.~Hu, R.~R. Selvaraju, C.~Ramaiah, R.~Xu, J.~F. JaJa, and
  L.~S. Davis, ``Tag: Boosting text-vqa via text-aware visual question-answer
  generation,'' {\em arXiv preprint arXiv:2208.01813}, 2022.

\bibitem{devlin2018bert}
J.~Devlin, M.-W. Chang, K.~Lee, and K.~Toutanova, ``Bert: Pre-training of deep
  bidirectional transformers for language understanding,'' {\em arXiv preprint
  arXiv:1810.04805}, 2018.

\bibitem{liu2019omnidirectional}
Y.~Liu, S.~Zhang, L.~Jin, L.~Xie, Y.~Wu, and Z.~Wang, ``Omnidirectional scene
  text detection with sequential-free box discretization,'' {\em arXiv preprint
  arXiv:1906.02371}, 2019.

\bibitem{wang2019simple}
P.~Wang, L.~Yang, H.~Li, Y.~Deng, C.~Shen, and Y.~Zhang, ``A simple and robust
  convolutional-attention network for irregular text recognition,'' {\em arXiv
  preprint arXiv:1904.01375}, vol.~6, no.~2, p.~1, 2019.

\bibitem{ren2015faster}
S.~Ren, K.~He, R.~Girshick, and J.~Sun, ``Faster r-cnn: Towards real-time
  object detection with region proposal networks,'' {\em Advances in neural
  information processing systems}, vol.~28, 2015.

\bibitem{lu2021localize}
X.~Lu, Z.~Fan, Y.~Wang, J.~Oh, and C.~P. Ros{\'e}, ``Localize, group, and
  select: Boosting textvqa by scene text modeling,'' in {\em Proceedings of the
  IEEE/CVF International Conference on Computer Vision}, pp.~2631--2639, 2021.

\bibitem{liu2020cascade}
F.~Liu, G.~Xu, Q.~Wu, Q.~Du, W.~Jia, and M.~Tan, ``Cascade reasoning network
  for text-based visual question answering,'' in {\em Proceedings of the 28th
  ACM International Conference on Multimedia}, pp.~4060--4069, 2020.

\bibitem{zhang2021position}
X.~Zhang and Q.~Yang, ``Position-augmented transformers with entity-aligned
  mesh for textvqa,'' in {\em Proceedings of the 29th ACM International
  Conference on Multimedia}, pp.~2519--2528, 2021.

\bibitem{sidorov2020textcaps}
O.~Sidorov, R.~Hu, M.~Rohrbach, and A.~Singh, ``Textcaps: a dataset for image
  captioning with reading comprehension,'' in {\em European Conference on
  Computer Vision}, pp.~742--758, Springer, 2020.

\bibitem{xie2017aggregated}
S.~Xie, R.~Girshick, P.~Doll{\'a}r, Z.~Tu, and K.~He, ``Aggregated residual
  transformations for deep neural networks,'' in {\em Proceedings of the IEEE
  conference on computer vision and pattern recognition}, pp.~1492--1500, 2017.

\bibitem{kuznetsova2020open}
A.~Kuznetsova, H.~Rom, N.~Alldrin, J.~Uijlings, I.~Krasin, J.~Pont-Tuset,
  S.~Kamali, S.~Popov, M.~Malloci, A.~Kolesnikov, {\em et~al.}, ``The open
  images dataset v4,'' {\em International Journal of Computer Vision},
  vol.~128, no.~7, pp.~1956--1981, 2020.

\bibitem{wang2021confidence}
Z.~Wang, R.~Bao, Q.~Wu, and S.~Liu, ``Confidence-aware non-repetitive
  multimodal transformers for textcaps,'' in {\em AAAI}, 2021.

\bibitem{anderson2018bottom}
P.~Anderson, X.~He, C.~Buehler, D.~Teney, M.~Johnson, S.~Gould, and L.~Zhang,
  ``Bottom-up and top-down attention for image captioning and visual question
  answering,'' in {\em Proceedings of the IEEE conference on computer vision
  and pattern recognition}, pp.~6077--6086, 2018.

\bibitem{UIT}
K.~Nguyen, ``Uit together research group,'' 2022.

\bibitem{huang2019attention}
L.~Huang, W.~Wang, J.~Chen, and X.-Y. Wei, ``Attention on attention for image
  captioning,'' in {\em Proceedings of the IEEE/CVF International Conference on
  Computer Vision}, pp.~4634--4643, 2019.

\bibitem{papineni2002bleu}
K.~Papineni, S.~Roukos, T.~Ward, and W.-J. Zhu, ``Bleu: a method for automatic
  evaluation of machine translation,'' in {\em Proceedings of the 40th annual
  meeting of the Association for Computational Linguistics}, pp.~311--318,
  2002.

\bibitem{denkowski2014meteor}
M.~Denkowski and A.~Lavie, ``Meteor universal: Language specific translation
  evaluation for any target language,'' in {\em Proceedings of the ninth
  workshop on statistical machine translation}, pp.~376--380, 2014.

\bibitem{rouge2004package}
L.~C. ROUGE, ``A package for automatic evaluation of summaries,'' in {\em
  Proceedings of Workshop on Text Summarization of ACL, Spain}, 2004.

\bibitem{anderson2016spice}
P.~Anderson, B.~Fernando, M.~Johnson, and S.~Gould, ``Spice: Semantic
  propositional image caption evaluation,'' in {\em European conference on
  computer vision}, pp.~382--398, Springer, 2016.

\bibitem{vedantam2015cider}
R.~Vedantam, C.~Lawrence~Zitnick, and D.~Parikh, ``Cider: Consensus-based image
  description evaluation,'' in {\em Proceedings of the IEEE conference on
  computer vision and pattern recognition}, pp.~4566--4575, 2015.

\end{thebibliography}

\end{document}